\documentclass{article} 
\usepackage{iclr2025_conference,times}

\usepackage{amsmath,amsfonts,bm}

\def\eqref#1{equation~\ref{#1}}

\def\1{\bm{1}}

\DeclareMathAlphabet{\mathsfit}{\encodingdefault}{\sfdefault}{m}{sl}
\SetMathAlphabet{\mathsfit}{bold}{\encodingdefault}{\sfdefault}{bx}{n}

\newcommand{\softmax}{\mathrm{softmax}}

\usepackage{hyperref}
\usepackage{url}
\usepackage{graphicx} 
\usepackage{subfigure}
\usepackage{relsize}
\usepackage{tikz}
\usepackage{pgfplots}
\usepackage{comment}
\usepackage[export]{adjustbox}
\usepackage{placeins}  
\usepackage{float}     

\usepackage[utf8]{inputenc} %
\usepackage[T1]{fontenc}    %
\usepackage{hyperref}       %
\usepackage{url}            %
\usepackage{booktabs}       %
\usepackage{amsfonts}       %
\usepackage{nicefrac}       %
\usepackage{microtype}      %
\usepackage{xcolor}         %
\usepackage{amsmath,amsthm,amssymb}
\usepackage[toc,page]{appendix}
\usepackage{algorithm}
\usepackage[noend]{algpseudocode}

\usepackage{subcaption}
\usepackage{multirow}

\usepackage{xspace}
\usepackage{wrapfig}

\usepackage[inline]{enumitem}

\usepackage{graphicx}
\usepackage{grffile}
\usepackage{color}

\usepackage{newtxmath}
\usepackage{listings}
\usepackage{color}

\usepackage[compact]{titlesec} 

\definecolor{dkgreen}{rgb}{0,0.6,0}
\definecolor{gray}{rgb}{0.5,0.5,0.5}
\definecolor{mauve}{rgb}{0.58,0,0.82}

\newtheorem{thm}{Theorem}
\newtheorem{obs}{Observation}
\title{Tree Attention: Topology-Aware Decoding for Long-Context Attention on GPU Clusters}

\iclrfinalcopy
\author{Vasudev Shyam$^{1}$\thanks{Joint first-authors}$\:$, Jonathan Pilault$^{1*}$, Emily Shepperd$^{2}$, Quentin Anthony$^{1}$, Beren Millidge$^{1}$ \\ 
$^1$Zyphra, $^2$2EleutherAI\\
\texttt{jonathan.pilault@gmail.com, vasu@zyphra.com}}

\begin{document}

\maketitle

\begin{abstract}
Self-attention is the core mathematical operation of modern transformer architectures and is also a significant computational bottleneck due to its quadratic complexity in the sequence length. 
In this work, we derive the scalar energy function whose gradient computes the self-attention block, thus elucidating the theoretical underpinnings of self-attention. 
Our formulation reveals that the reduction across the sequence axis can be efficiently computed in parallel through a tree reduction. 
Our algorithm, called \texttt{Tree Attention}, for parallelizing exact attention computation across multiple GPUs enables cross-device decoding to be performed \emph{asymptotically} faster (up to $8\times$ faster in our experiments) than state-of-the-art approaches such as \texttt{Ring Attention}, while also requiring significantly less communication volume and incurring $2\times$ less peak memory. 
We demonstrate that \texttt{Tree Attention} speeds up decoding up to 4x on Llama 3.1-8B and can be applied to a variety of hardware and networking setups such as H100 DGX nodes, AMD MI300x nodes, and PCIe connected NVIDIA RTX 4090s.
Our code is publicly available here: \url{https://github.com/Zyphra/tree_attention}

\end{abstract}

\section{Introduction}

The self-attention operation is the core computational building block of the transformer architecture \cite{bahdanau2014neural,vaswani2017attention}, which has become an ubiquitous and highly effective workhorse architecture currently applied at scale to language \cite{brown2020language,kaplan2020scaling,hoffmann2022training,team2023gemini,achiam2023gpt,ijcai2023p460}, vision \cite{dosovitskiy2020image}, audio \cite{betker2023better}, and decision-making \cite{chen2021decision,reed2022generalist}.
Nonetheless, the quadratic time complexity of self-attention means that significant resources are required to train and generate from transformer-based Large Language Models (LLMs), especially for models with large context lengths.

During inference, the attention block largely determines the computational and memory requirements, which become more demanding as the input sequence length increases.
Although LLMs generate one token at a time, the entire sequence of past tokens must still be stored in memory and used to compute attention scores during generation.
Since attention performs a similarity matching of every token representation with every other, it incurs quadratic computational complexity in terms of flops.

There have been recent advances in training LLMs to handle extremely long contexts (up to 1M tokens) ~\cite{chen2023extending,kaiokendev,peng2023yarn}.
Such models attain qualitatively new capabilities such as extremely large-scale in-context learning of entire small datasets held in the prompt \cite{reid2024gemini,lee2024can,bertsch2024context}. 
They can also avoid putting multi-modal continuous data through a lossy tokenization scheme \cite{reid2024gemini,team2024chameleon} by directly operating at the byte level \cite{xue2022byt5,wu2024beyond}. 
The issue however is that performing inference on such long contexts is very expensive.

To speed up inference and alleviate memory requirements, recent works have attempted to alter the attention mechanism itself, either by linearizing it \cite{katharopoulos2020transformers}, or approximating it by a kernel map \cite{choromanski2020rethinking,peng2021random,arora2024simple}, which reduces the complexity to linear at the cost of reduced expressiveness. 
Others have invented alternative sequence mixing architectures such as state-space models which are designed to be efficiently computable in linear time and constant memory \cite{gu2023mamba,dao2024transformers,katsch2023gateloop,sun2023retentive,glorioso2024zamba}.

\begin{figure*}[ht]
  \begin{center}
      \mbox {
          \hspace{-0.03\linewidth}
           \subfigure[Multi Node \texttt{Tree Attention} (Ours)]
          {
            \includegraphics[width=.48\linewidth]{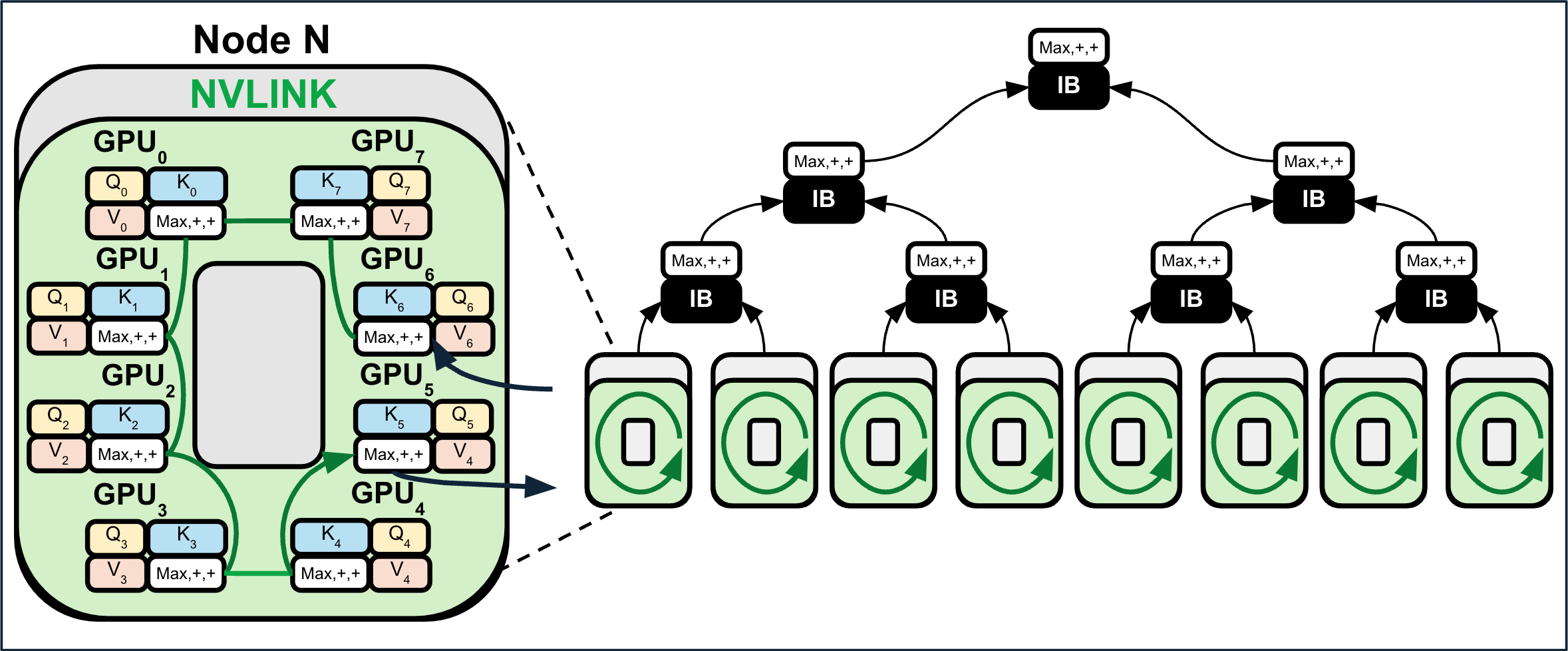}
            \label{fig:tree}
          }
          \hspace{-0.023\linewidth}
           \subfigure[Multi Node \texttt{Ring Attention}]
          {
            \raisebox{0mm}{\includegraphics[width=.408\linewidth]{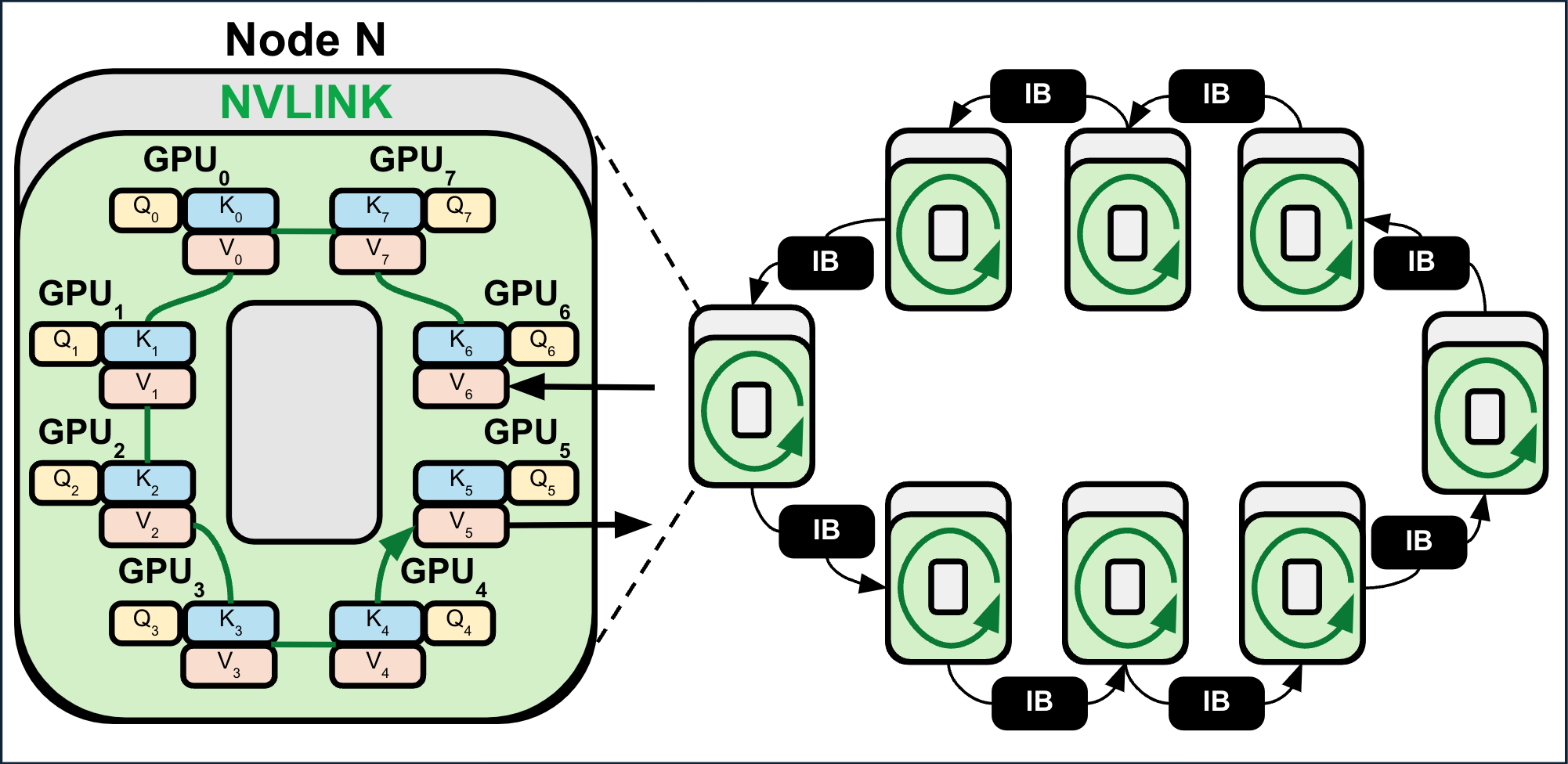}}
            \label{fig:ring}
          }
          \hspace{-0.023\linewidth}
          {
            \includegraphics[width=.15\linewidth]{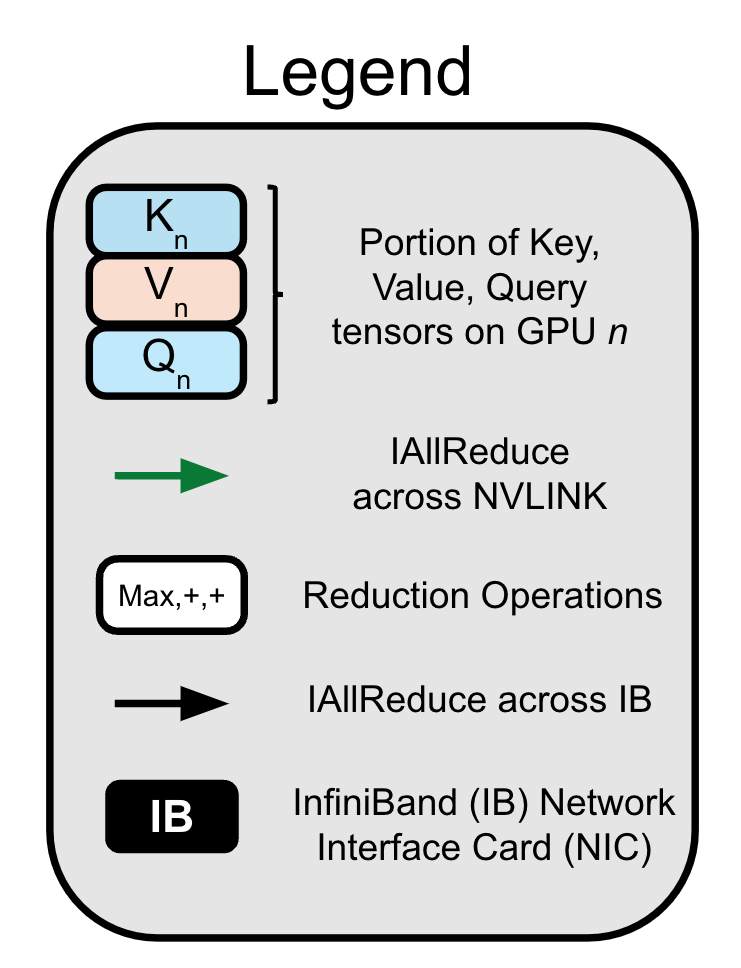}
          }
      }
      \vspace*{-0.5\baselineskip}
      \caption{\small Ring and \texttt{Tree Attention} Topologies. Due to the associative properties of the logsumexp and max operations of \texttt{Tree Attention} (Fig.~\ref{fig:tree}), is possible to structure the reduction across the sequence as a tree, requiring asymptotically fewer communication steps than \texttt{Ring Attention} (Fig.~\ref{fig:ring}) as well as less memory and communications volume.}
      \label{fig:topologies}
  \end{center}
\vspace*{-1\baselineskip}
\end{figure*}

Other approaches utilize efficient algorithms to reduce the computational burden of attention while keeping the core computation the same. 
These include memory-efficient attention~\cite{mem_attn}, \texttt{Flash Attention} \cite{dao2022flashattention} and \texttt{Flash Decoding} \cite{flashdecoding}, which provide a set of IO-aware kernels to map the attention operation to the GPU hardware resources in an extremely efficient way, significantly reducing the memory overhead required.
Further works~\cite{CharacterAI2024, kang2024gearefficientkvcache, liu2024minicachekvcachecompression, nawrot2024dynamicmemorycompressionretrofitting} explore compressing or otherwise reducing the KV cache required in generation.
Finally, \texttt{Ring Attention} \cite{Ring_Attn} proposes a way to parallelize the attention computation across the sequence axis between GPUs, thus enabling significantly longer contexts than can be served on a single GPU. 
Since our proposed method is an exact calculation of attention\footnote{It can be shown empirically that Ring Attention and Tree Attention are exact computations of Attention since both methods have exactly the same activations as the forward pass of Vanilla Attention.}, it is a plugin replacement for any multi-GPU sequence parallel mechanism such as the state of the art \texttt{Ring Attention} mechanisms.
By leveraging the exact energy
function for the self-attention block, we develop a method to speed up inference for long context use-cases when keys and values are sharded across multiple GPUs along the sequence axis.

Our proposed algorithm for computing attention via the gradient of the energy function is built on top of an efficient parallel computation and tree reduction communication strategy.
In particular, this formulation lets us devise an asymptotically faster algorithm for performing decoding in which the number of communication steps scales logarithmically with the number of devices, instead of linearly in alternatives such as \texttt{Ring Attention} \cite{Ring_Attn}. 
Our topology-aware approach illustrated in Fig.~\ref{fig:topologies} significantly outperforms leading attention parallelization methods such as \texttt{Ring Attention} on multiple devices.

\section{Related Works}

The computational complexity of self-attention, introduced by \cite{vaswani2017attention}, poses challenges for long sequences due to its quadratic dependency on sequence length, $O(n^2 \cdot d)$. 
To address this, attention \textbf{approximation} mechanisms like Linformer \citep{linformer} and Performer \citep{performer} reduce complexity to linear $O(n)$ using low-rank projections and kernelized approximations on a \emph{single device}.
Sparse models such as Longformer \citep{longformer} and BigBird \citep{bigbird} further optimize computations by restricting attention to local windows or sparsity patterns, significantly reducing resource demands while maintaining performance for specific tasks.
Such methods however provide approximations to the attention mechanism while we seek to parallelize the \textbf{exact} attention computation across the sequence axis.

Theoretical work has also contributed to improving the efficiency of both exact and approximate methods. Kernel-based approaches, such as those by \cite{DBLP:journals/corr/abs-1908-11775}, suggest alternative formulations to self-attention that are computationally efficient. 
Surveys like \cite{DBLP:journals/corr/abs-2009-06732} highlight these advancements, emphasizing the synergy between parallelization strategies and sparsity or approximation techniques, ensuring self-attention remains scalable even in constrained computational environments.
It must be noted as well that \cite{Keles} established lower bounds on the computational complexity of self-attention, demonstrating that achieving sub-quadratic time complexity is unlikely unless the Strong Exponential Time Hypothesis (SETH) is false.

In addition to approximation methods, several approaches focus on parallelizing the exact attention computations. 
\texttt{FlashAttention} \citep{dao2022flashattention}, for instance, reorganizes the attention computation into smaller, memory-efficient blocks that leverage GPU memory hierarchies to enable faster and parallelized processing of exact attention, and by doing so reduces the memory complexity from quadratic to linear. 
Other techniques use optimized matrix operations and tiling strategies to distribute attention computations across cores or threads efficiently \citep{shen2021efficient}. 
While these methods aim to maximize throughput while maintaining the precision of exact attention, they focus on speeding up single-device attention computation.
Since we parallelize exact attention across multiple devices, \texttt{Ring Attention} \citep{Ring_Attn} is most comparable to our work.
Finally, to the best of our knowledge, there are no other techniques that explore multi-device parallel decoding as we have done in this paper.

\section{Self-Attention}

The self-attention operation can be represented as a set of dot product similarity searches between queries and keys. 
These similarity scores are then reduced along the sequence axis and softmaxed, so that for a given query, there is a probability distribution of the similarities of each given key. We then take the expectation of the value vectors against this distribution.
We denote the queries assigned to a sequence of length $N$ as $\{q_a,a=1,\cdots N\}$, where each query is a vector of size $d$ that stands for hidden dimension, $q_a\in \mathbb{R}^{d}$, and similarly the keys and values $\{(k_a,v_a),a=1,\cdots N\}$.
Attention can be written as
\begin{align*}
    z^a =  \sum^N_{i=1}\softmax(q_a\cdot k_i^T) v_i\,.
\end{align*}

Naively computing attention in this way requires materializing the $qk$ matrix with computational and memory cost quadratic in the sequence length.
Memory-efficient attention \cite{mem_attn} is an iterative way to compute the softmax similarities without ever having to materialize the full attention matrix. 
It performs the following operations, one query (or a chunk of queries) at a time:
\begin{align}
    s^{(j)}_i &= \text{exp}(q_j\cdot k_i) \\
    n^{(j)}_i &= n^{(j)}_{i-1} + v_i s^{(j)}_i \\
    d^{(j)}_i &= d^{(j)}_{i-1} + s^{(j)}_i
\end{align}
Then, once the values $v$ and softmax denominator $d$ are computed, we divide to get the final softmaxed scores $z^{(j)} = \frac{n^{(j)}}{d^{(j)}}$ for every query index $j$.
Computing attention in this iterative manner significantly reduces the required memory.

\texttt{Flash Attention} \cite{dao2022flashattention} utilizes a similar approach to reduce the memory and computational cost of attention, but the algorithm is not adapted for multi-GPU computation.
\texttt{Flash Attention} performs the iterative algorithm of \cite{mem_attn} in a blockwise manner, utilizing the block-parallel computational primitives available inside single GPU tensor cores.
Additionally, it precisely sizes the blocks such that they can fit into the SRAM of the GPU for the entire attention computation, effectively performing kernel fusion and preventing many unnecessary IO operations.

\section{Self-Attention as the Gradient of an Energy Function}
\label{sec:attn_grad}

Following the ubiquitous success of the transformer architecture, there has been significant effort to mathematically understand the nature and meaning of the attention operation and link it to energy models \citep{krotov2016dense,HAM,millidge2022universal,hoover2024dense}, such as Hopfield Networks \citep{Hopfield,damico2024selfattentionattractornetworktransient}.
\citet{Hopfield} pioneered this field by performing a similar but distinct analysis to relate self-attention with the modern Hopfield networks, providing a novel and insightful interpretation of self-attention as performing hetero-associative memory lookups using a high-powered nonlinear similarity function. 
This work was later extended by \citet{energy_transformer}, who derived a modified version of the transformer based off an energy function.
However, while it has long been known that the softmax operation can be derived as the gradient of the following scalar function:
\begin{equation}
\partial_{z_j}\log \sum^n_{a=1} \exp(z_a)=\frac{e^{z_j}}{\sum^n_{a=1}e^{z_a}}=\textrm{softmax}(z_j), 
\end{equation}
known as the log-sum-exp, an equivalent function for the self-attention block has not yet been derived. 
We develop in this paper a link between attention and energy functions by introducing an auxiliary \textit{source} vector $\zeta$, which represents the ``external contributions'' to the system's energy \citep{energy_physical_sys_1982}.
The \textit{source} $\zeta$ is the parameter with respect to which we compute the gradient of the scalar energy function to obtain the self-attention operation.
As we will see, we need the source in order to write down the generating function of the moments of the distribution since taking the gradient with respect to $\zeta$ yields the exact self-attention operation.

This insight allows us to make the following observation:

\begin{obs}
\label{obs:1}
Attention can be expressed at the gradient of an scalar energy function $F(\zeta)$ with respect to the \textit{source} $\zeta$, such that:

\begin{equation}
\sum^N_{a=1}\softmax(q\cdot k_a)v_a = \frac{\partial F}{\partial \zeta}\bigg\vert_{\zeta=0},
\end{equation}

where the moment generating function (i.e. the energy function) $F(\zeta)$ is defined as:

\begin{equation}
F(\zeta) = \log \sum_a\exp\left(q\cdot k_a^T +\zeta\cdot v^T_a\right).
\end{equation}

\end{obs}

The proof of Observation~\ref{obs:1} can be found in Appendix~\ref{app:proof_attn_energy}.
Please note that this formulation also allows to make a Bayesian interpretation of Attention in Appendix~\ref{app:bayesian} and motivates our \textrm{Tree Attention} algorithm in the next Section~\ref{sec:TreeAttn}.

\section{Tree Attention}
\label{sec:TreeAttn}
In this section we show how the formulation of the attention operation as the gradient of an energy function suggests an efficient parallel strategy for computing it. The key insight is to leverage an efficient algorithm to compute the energy, and then differentiate it in order to obtain an efficient algorithm to compute attention. 

\subsection{Efficient Energy Function Computation}

Let us focus on the case of decoding with a KV cache in a causal language model where we have one query and $N$ keys and values. In this case, the energy function is:
\begin{eqnarray}
F_{dec} = \log\sum^N_{a=1}\exp(q\cdot k_a^T+\zeta\cdot v^T_a)
\equiv \textrm{logsumexp}_a(\{q\cdot k^T_a+\zeta\cdot v^T_a,a=1,\cdots,N\}).
\end{eqnarray}
A crucial fact is that both \(\textrm{logsumexp}_a\) and \(\max_a\) are associative operations:
        \begin{eqnarray*}
        \textrm{logsumexp}_a(\{T_a, \textrm{logsumexp}_a(\{R_a, S_a\})\}) =
        \textrm{logsumexp}_a(\{\textrm{logsumexp}_a(\{T_a, R_a\}), S_a\}),
         \end{eqnarray*}
         \begin{eqnarray*}
        \max_a(\{\max_a(\{T_a, R_a\}), S_a\}) =\max_a(\{T_a, \max_a(\{R_a, S_a\})\}).
        \end{eqnarray*}

We can prove that this associative property allows these reductions to be performed efficiently in parallel with logarithmic time complexity, provided we have adequately many parallel workers: 

\begin{thm}
\label{thm:1}
The time complexity of a reduction operation involving an associative function, such as \(\textrm{logsumexp}_a\) or \(\max_a\), over an array of size \(N\) using \(p\) parallel processors is \(O\left(\frac{N}{p} + \log p\right)\). 
When the number of processors \(p\) is equal to \(N\), the time complexity is reduced to \(O(\log N)\).
\end{thm}
The proof of Theorem~\ref{thm:1} is in Appendix~\ref{app:proof}.

Putting this result together, and for $\hat{a},\hat{b} \in \{1,\cdots,t\}$ intra-chunk indices, we get the following highly parallel Algorithm~\ref{alg:energy_fwd}:

\begin{algorithm}[H]
\caption{Single Query Energy Forward (calculating logsumexp)}
\begin{algorithmic}[1]
\item Divide $\mathbf{k},\mathbf{v}\in \mathbb{R}^{N\times d_h}$ into $p$ chunks $\{\mathbf{k}_{\hat{a}},\mathbf{v}_{\hat{a}},\hat{a}\in \{1,\cdots,N/p\}\}$ of size $t = N/p$
\item Scatter a copy of $\mathbf{q},\mathbf{\zeta}$, and each $\mathbf{k}_{\hat{a}}, \mathbf{v}_{\hat{a}}$ to each of the $p$ processors. 
\item In parallel compute $r_{\hat{a}} = \mathbf{q}\cdot \mathbf{k}^T_{\hat{a}}+\mathbf{\zeta}\cdot \mathbf{v}^T_{\hat{a}}$
\item Compute $m = \textrm{Reduce}(\max,r_{\hat{a}})$ by doing a tree reduction.
\item Scatter $m$ to every device and update $r_{\hat{a}}\rightarrow r_{\hat{a}}-m$.
\item Compute $lse = \textrm{Reduce}(\textrm{logsumexp},r_{\hat{a}})$ by doing a tree reduction.
\item Save $lse,m$ for gradient w.r.t $\zeta$.
\item Return $lse$
\end{algorithmic}
\label{alg:energy_fwd}
\end{algorithm}

\subsection{Efficient parallel decoding}

One of the core insights of automatic differentiation is that the gradient of a function $\nabla_x f(x)$ can be computed with the same time complexity as computing $f(x)$ \cite{f_v_gf}.
The caveat however is that if the function has a deep computational graph, then the memory footprint of computing the gradient grows with that depth as backpropagation requires storing the values of the intermediate tensors.
In our case, the computational graph involved in computing the energy is shallow and therefore the memory overhead is negligible.
This means that if we can compute the energy efficiently, we obtain an efficient algorithm for computing its gradient (i.e. the self-attention operation) automatically.

In our case, we want to compute the gradient of the energy function with respect to $\zeta_A$ and then set it to zero. 
This can be done with automatic differentiation engines having set $\zeta$ to be a tensor of zeros from the very outset. 
We can however manually implement a gradient with respect to $\zeta$ pass of the above Algorithm ~\ref{alg:energy_fwd} that does not materialize $\zeta$ in Algorithm~\ref{alg:zet_grad} below. 
Note in particular that when we set $\zeta_A=0,\,\, A\in \{1,\cdots, d_h\}$ then $lse$ involves only the logsumexp of the dot product between queries and keys.

\begin{algorithm}[H]
\caption{\texttt{Tree Decoding} (using atomic operation on each device)}
\begin{algorithmic}[1]
\item Divide $\mathbf{k},\mathbf{v}\in \mathbb{R}^{N\times d_h}$ into $p$ chunks $\{\mathbf{k}_{\hat{a}},\mathbf{v}_{\hat{a}},\hat{a}\in \{1,\cdots,N/p\}\}$ of size $t = N/p$
\item Calculate $m$ and $lse$ using Algorithm ~\ref{alg:energy_fwd}.
\item Scatter a copy of $\mathbf{q},m$ and $lse$, and each $\mathbf{k}_{\hat{a}}, \mathbf{v}_{\hat{a}}$ to each of the $p$ processors. 
\item In parallel compute $r_{\hat{a}} = \mathbf{q}\cdot \mathbf{k}^T_{\hat{a}}-m$
\item Compute $R_{\hat{a}}=\frac{\exp(r_{\hat{a}})}{\exp(lse)} \cdot v_{\hat{a}} = \exp(r_{\hat{a}}-lse)\cdot v_{\hat{a}}$
\item Compute $z = \textrm{Reduce}(\textrm{sum},R_{\hat{a}})$
\item Return $z$
\end{algorithmic}
\label{alg:zet_grad}
\end{algorithm}
Notice here that by storing $lse,m$ for the backward pass, the only remaining reduction operation that needs to be performed is the one in line 5 of the above algorithm. This single reduction takes $O(N/p)$ time to compute the local sums on each device and $\log p$ time to communicate and combine partial results, and therefore we get the same asymptotic complexity as the logsumexp calculation. 

In practice, we implement the forward and gradient w.r.t. $\zeta$ in a single function which returns both the value and the gradient of the energy function. 
We can therefore put together Algorithms \ref{alg:energy_fwd} and \ref{alg:zet_grad} into the following efficient parallel decoding Algorithm \ref{alg:tree_dec}:

\begin{algorithm}[H]
\caption{\texttt{Tree Decoding} (using \texttt{Flash Attention 2} on each device)}
\label{alg:tree_dec}
\begin{algorithmic}[1]
\item Divide $\mathbf{k},\mathbf{v}\in \mathbb{R}^{N\times d_h}$ among $p$ GPUs, each with a chunk $\{\mathbf{k}_{\hat{a}},\mathbf{v}_{\hat{a}},\hat{a} \in \{1,\cdots, N/p\}\}$ of size $t = N/p$ and scatter $\mathbf{q}$ to each GPU.
\item Use \texttt{Flash Attention 2} to compute $\textrm{o}=\frac{\sum_{\hat{a}}\exp(\mathbf{q}\cdot \mathbf{k}_{\hat{a}}^T)\mathbf{v}_{\hat{a}}}{\sum_{\hat{b}}\exp(\mathbf{q}\cdot \mathbf{k}^T_{\hat{b}})}$ and $\textrm{lse}=\log\sum_{\hat{b}}\exp(\mathbf{q}\cdot \mathbf{k}^T_{\hat{b}})$.

\item Recompute the global max $m = \textrm{\texttt{Allreduce}}(\textrm{max},\textrm{lse})$.
\item Get local numerator and denominator by computing: $\textrm{n} = \textrm{o}*\exp(\textrm{lse}-m),\textrm{d} = \exp(\textrm{lse}-m)$.
\item Compute global numerator and denominator with: $\textrm{n}_g =\textrm{\texttt{Allreduce}}(\textrm{sum},\textrm{n}) ,\textrm{d}_g = \textrm{\texttt{Allreduce}}(\textrm{sum},\textrm{d})$. 
\item Return result $z = \frac{\textrm{n}_g}{\textrm{d}_g}$.
\end{algorithmic}
\end{algorithm}

This algorithm requires three \texttt{Allreduce} operations in total, meaning that the required time complexity is $O(3(N/p+\log p ))$.

\subsection{Efficient collective operations using topology-awareness}
\paragraph{Communication overheads}
While the theoretical analysis above indicates that we should see speedups when using tree-based reductions, this is not necessarily guaranteed in practice due to various potential overheads. 
In particular, our argument for the time complexity of our proposed \texttt{Tree Decoding} algorithm assumes that communication of partial results is instantaneous, which in practice is never the case. 
In fact, as we scale the sequence length, or the number of GPUs especially to the multi-node setting, the time taken for communication is the dominant contribution to the total execution time. 
However, importantly, beyond its asymptotic benefits, \texttt{Tree Attention} benefits from taking advantage of the two-level topology which is standard in modern GPU clusters.

\begin{wrapfigure}{r}{0.55\linewidth}
\vspace{-2ex}
    \centering
    \includegraphics[width=1\linewidth]{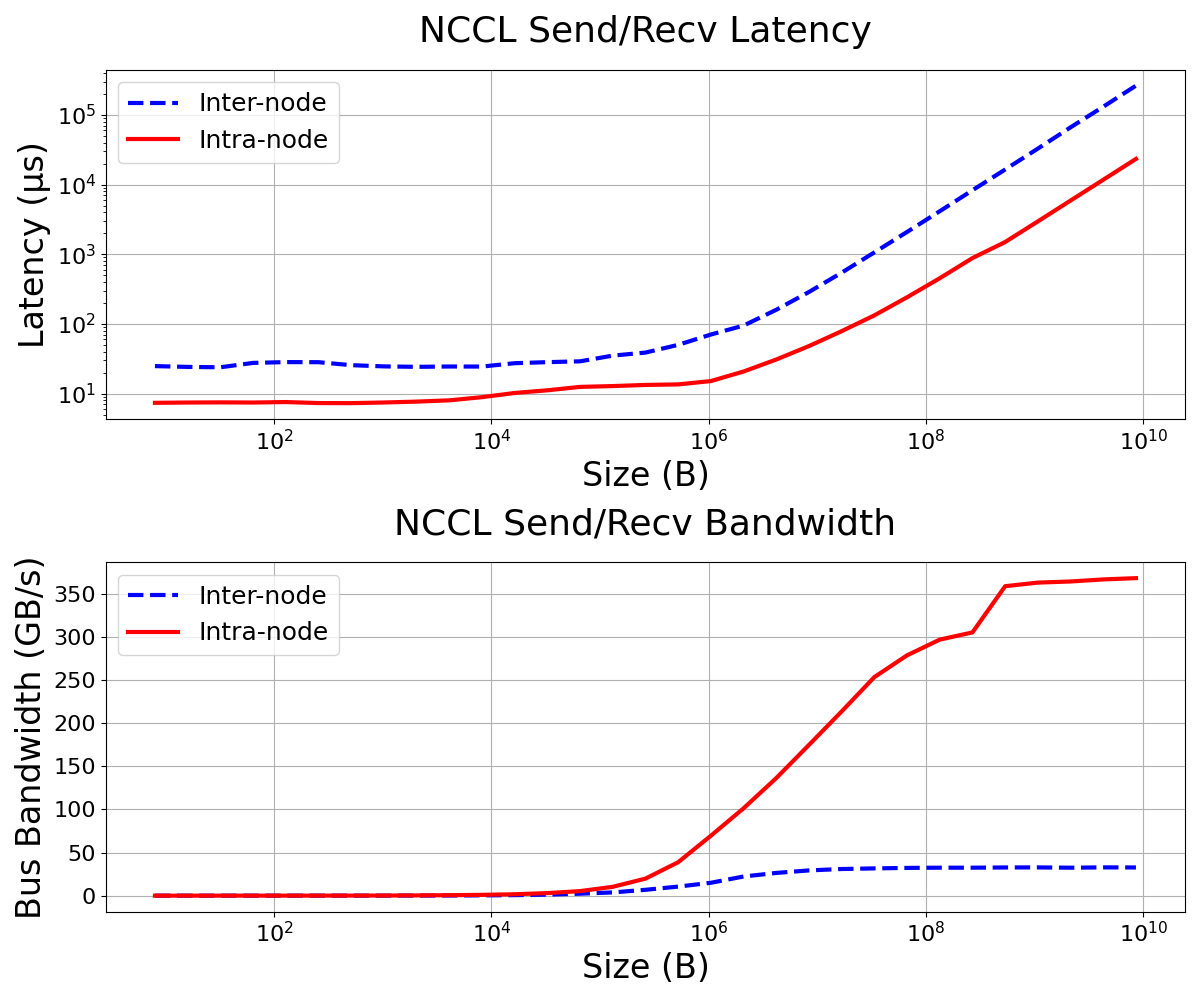}

    \caption{\small NCCL Send/Recv between two H100 GPUs intra-node and inter-node. GPU clusters offer a two-tier topology where intra-node bandwidth is significantly higher than inter-node. Algorithms such as \texttt{Tree Attention} exploit this topology by reducing inter-node communication requirements, enabling better overlap of communication with computation.}
    \label{fig:nccl-pt2pt}
\end{wrapfigure}

We benchmark our algorithm against a previously proposed sequence parallel attention algorithm called \texttt{Ring Attention}. 
Like our algorithm, \texttt{Ring Attention} assumes that the sequence is sharded across GPUs and performs the attention computation without gathering all of the sequence on to a single device. 
Instead, it communicates shards of the keys and values in a point-to-point manner between neighboring GPUs that are logically arranged in a ring topology. This communication is overlapped with the computation of the local shard of the output. 
In contrast with this strategy, our algorithm scatters the query and communicates the partial result across all GPUs when performing the \texttt{AllReduce} operation, but does not move the key and value shards between GPUs. Consequently, in the decoding case, our method benefits from having lower communication volume and suffers less from the communication cost overhead than \texttt{Ring Attention} does.

\paragraph{Implications of network bandwidth heirarchy}
\texttt{Ring Attention} is inherently not topology-aware, and only scales within a network of homogeneous bandwidth. 
However, this is in conflict with the two-level network topology of modern GPU clusters, which use high-bandwidth interconnects within nodes (NVLINK or PCIe) and comparatively lower-bandwidth interconnects across nodes (InfiniBand or Ethernet). 
The interconnects greatly differ in bandwidth and latency (see Figure \ref{fig:nccl-pt2pt}). 
Therefore, \texttt{Ring Attention} is bottlenecked by the slowest interconnect, and cannot always overlap the attention computation with communication. We discuss this point further in \ref{sec:comm_vol}
\texttt{Tree Attention} improves on \texttt{Ring Attention} by using network topology-aware communication patterns to increase overlap of computation and communication, and decrease this scalability bottleneck on communication from the distributed attention computation.

In practice, collective communication libraries like NCCL attempt to automatically detect what the right communication strategy is based on considerations such as data volume and network topology. 
In DGX clusters, for collective operations within a node, ring reduce is performed whereas a tree reduction is performed across nodes. 
We see that therefore using built-in collective operations such as \texttt{Allreduce} leads to a better performance when decoding from long contexts across multiple GPUs than enforcing the \texttt{Ring Attention}'s point to point communication pattern. 
We show how the following strategy outperforms \texttt{Ring Attention} when decoding from very long contexts across multiple GPUs.

In our empirical experiments , we use \texttt{Flash Attention 2} \citep{dao2023flashattention} within each device, both for our algorithm and for \texttt{Ring Attention}\footnote{A JAX-based \texttt{Ring Attention} implementation that uses \texttt{Flash Attention 2} can be found here: \url{https://github.com/nshepperd/flash_attn_jax}.}.
We provide a simple JAX implementation of our method in Appendix~\ref{app:code}. Note that our method mirrors \texttt{Flash Decoding} \citep{flashdecoding} except in that case, the parallelization happens at the level of different streaming multiprocessors (SMs) within a GPU whereas we parallelize between different GPUs.
All computations are performed in BF16.

\section{Results}
\label{sec:results}

Similar to \texttt{Ring Attention}, \texttt{Tree Attention} is an exact computation of attention.
Since training and evaluation metrics are the same as for attention, our experimental results are focused primarily on latency in section~\ref{sec:latency}, peak memory usage in section~\ref{sec:memory-cost} and communication volumes in section~\ref{sec:comm_vol}.
Since our algorithm computes numerically identical results as the forward pass of standard attention, our performance results transfer seamlessly to transformer architectures.

We performed experiments in Sections~\ref{sec:latency} to~\ref{sec:comm_vol} on a DGX H100 cluster consisting of 16 nodes, each containing 8 H100 GPUs. 
All GPUs within the node are connected via an all-to-all NVLINK 4.0 (900GBps) topology. 
Nodes are connected to each other via 8 InfiniBand NDR interconnects per node (1 per GPU), each of which provides 400 Gbps (leading to an aggregate 3.2 Tbps node injection bandwidth).

We also show \texttt{Ring Attention} and \texttt{Tree Attention} comparisons when used in a Llama 3 model \citep{grattafiori2024llama3herdmodels} in Sections~\ref{sec:llama_perf} and~\ref{app:llama_perf} on viarous GPU and interconnect types: 8 H100 GPUs with NVLINK 4.0, 8 AMD MI300X GPUs with AMD infinity fabric for intra-node communication and RoCE for inter-node communication, and 2 RTX 4090 GPUs with PCIe interconnect.

\subsection{Latency}
\label{sec:latency}

\begin{figure*}[htbp]
  \begin{center}
      \subfigure[Relative Execution time at different sequence lengths]
      {
        \raisebox{0mm}{\includegraphics[width=1\linewidth]{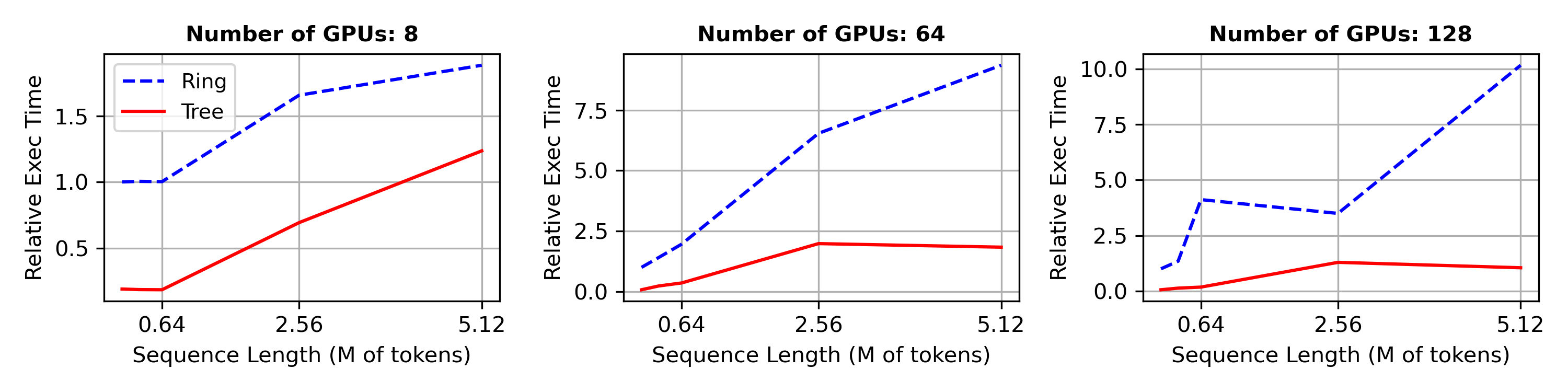}}
        \label{fig:relative_times}
      }
      \subfigure[Absolute Execution time for varying cluster sizes]
      {
        \includegraphics[width=1\linewidth]{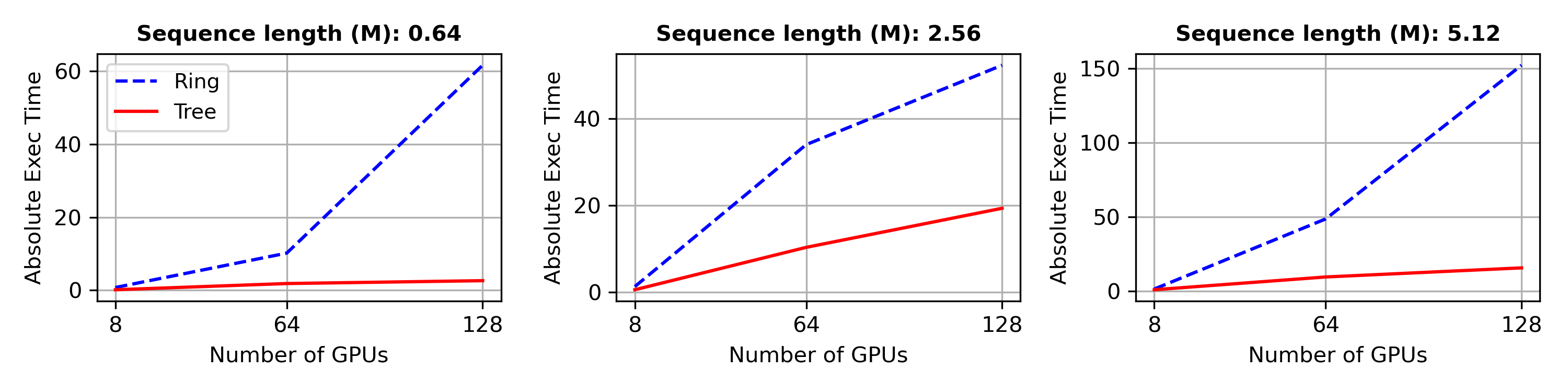}
        \label{fig:absolute_times}
      }
      \vspace*{-1.\baselineskip}
      \caption{\small Execution time of 16-head \texttt{Tree Attention} vs \texttt{Ring Attention} for different sizes of GPU cluster (from 1 to 16 H100 DGX nodes). Relative execution times are indexed to
the \texttt{Ring Attention} times at a sequence length of 80k tokens.}
      \label{fig:times}
  \end{center}
\vspace*{-1\baselineskip}
\end{figure*}

In terms of practical usefulness, our study of the energy function brought to light a previously unnoted parallelizability inside the attention computation -- that of the reduction of the logsumexp across the sequence dimension, which can be implemented as a parallel \texttt{Allreduce}.
As stated in Theorem~\ref{thm:1}, it becomes theoretically possible to implement attention, per query as an $N/p+ \log (p)$ parallel operations rather than $N$, where the logarithmic term is proportional to the number of devices available for parallelization.
When the attention is sharded across multiple devices, this asymptotic speedup creates a considerable speedup over alternative methods for decoding.

To empirically test the theoretical benefits of our \texttt{Tree Attention} method, we compute latency by measuring the time required to perform decoding for different sequence lengths and varying number of H100 nodes.
We compare \texttt{Tree Attention} to our own \texttt{Ring Attention} execution times in Fig.~\ref{fig:times}.
Both methods use \texttt{Flash Attention 2} \cite{dao2023flashattention} for the individual-GPU attention computation. 
For our experiments, we benchmark on a standard attention block consisting of 16 heads of dimension 128 across different sequence lengths.

Our latency results shows how \texttt{Tree Attention} improves over \texttt{Ring Attention} as we increase the sequence length in Fig.~\ref{fig:relative_times} and increase the number of GPUs in Fig.~\ref{fig:absolute_times}.
To better highlight execution time trends with an increasing sequence length, we have also added relative execution time of both methods with respect to the execution time of ring attention at a sequence length of 80k.
With relative execution time in Fig.~\ref{fig:relative_times}, we notice that Tree attention's execution time flattens as the number of GPUs increases, while Ring Attention relative execution time continues to increase.
As the plots demonstrate, as we scale the sequence length or the number of GPUs, the gap between \texttt{Tree Attention} and \texttt{Ring Attention} execution time widens \emph{asymptotically}.
Remarkably, \texttt{Tree Attention} achieves close $\times8$ speedups when we use 128 GPUs on a sequence length of 5.12M. 
We expect this trend to continue for larger sequence lengths.
Please note that our DGX cluster is made up of 16 nodes each with 8 GPUs.
Results for 8 GPUS use one node, for 64 GPUs uses 8 nodes and for 128 GPUs uses 16 nodes.

\subsection{Memory cost}
\label{sec:memory-cost}

\begin{figure}[ht]

    \centering
    \subfigure[Memory overhead by hidden size]{
        \includegraphics[width=0.48\linewidth]{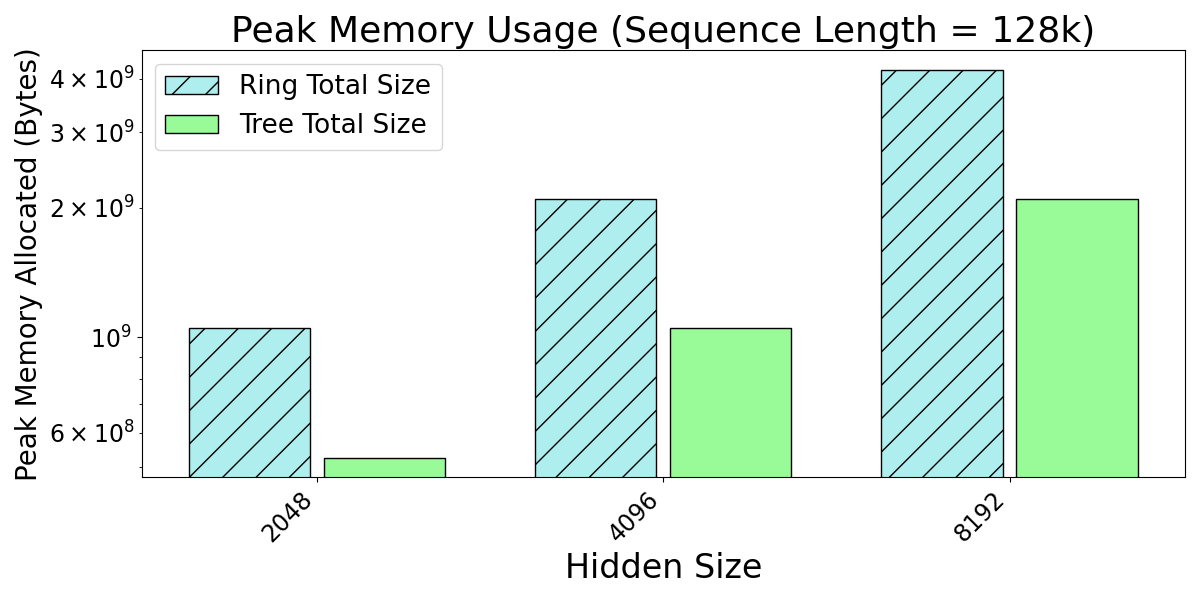}
        \label{fig:tree-mem-hidden}
    }
    \subfigure[Memory overhead by sequence length]{
        \includegraphics[width=0.48\linewidth]{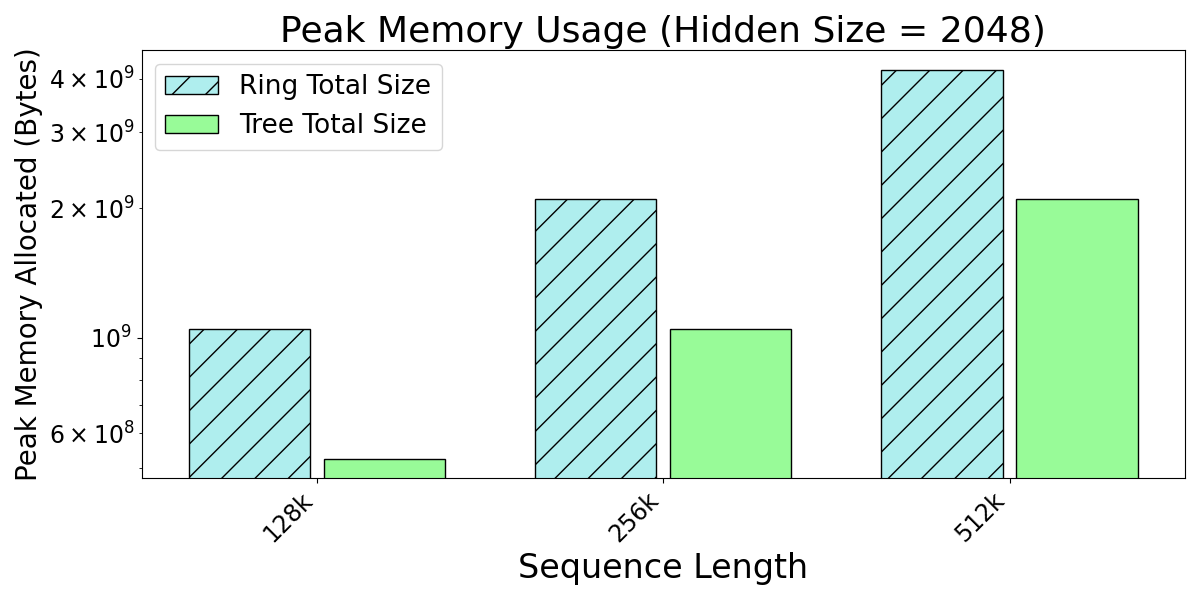}
        \label{fig:tree-mem-seqlen}
    }
    
    \caption{\small Peak memory usage of a single attention block with \texttt{Tree Attention} vs \texttt{Ring Attention} when sharded between two RTX 4090s. Results were taken using the JAX memory profiler on one GPU. The difference in peak memory scales with hidden size and sequence length.}
    \label{fig:mem}
\end{figure}

To perform \texttt{Ring Attention} with a distributed KV cache, it is necessary to broadcast the query corresponding to the final element of the sequence back to all devices, as outlined in step 2 of our Algorithm~\ref{alg:energy_fwd}.
Each device will then hold a tuple $(\mathbf{q}, \mathbf{k}_{\hat{a}}, \mathbf{v}_{\hat{a}})$, where $\hat{a}$ is the chunk index, which includes the query vector and a local chunk of the keys and values specific to the sequence chunk on that device. 
The memory cost to store these objects is the same as for \texttt{Tree Decoding}.  
Additionally, \texttt{Ring Attention} must store the $\mathbf{k}_{\hat{a}'
},\mathbf{v}_{\hat{a}'}$ coming from the neighbouring device and the chunk of the output $o$ that has the same shape as the query held by that device. 
In contrast, our method requires storing instead only the communicated chunk of the numerator $\textrm{n}$, denominator $\textrm{d}$ and max $m$. 
We do not pre-allocate an output tensor but instead just return the result of doing the \texttt{Allreduce} to the numerator divided by the Allreduced denominator. 
In summary we have the following peak memory costs for Ring and Tree attention: 

\begin{eqnarray}
\textrm{Mem}_{\text{ring}} &= 4btd+2bd \\
\textrm{Mem}_{\text{tree}} &= 2btd +2bd+2bn_h,
\end{eqnarray}

where $d = d_h\times n_h$, for head size $d_h$ and $n_h$ number of heads, $b$ denotes the batch size and $t=N/p$.
As such, so long as $2bn_h \leq 2btd$, which will almost always be the case in realistic scenarios, our method always has a lower peak memory cost compared to \texttt{Ring Attention}. 

We empirically measure peak memory utilization for our approach and \texttt{Ring Attention} to show that indeed memory usage is significantly less for \texttt{Tree Attention} in Figure \ref{fig:mem}.
As predicted by theory, scaling hidden size or sequence length scales \texttt{Ring Attention} peak memory usage about $2 \times$ faster than \texttt{Tree Attention}.
For example, doubling the hidden size from 2048 to 4096, doubles the gap in peak memory between two methods, going from 524MB to 1040MB.

\subsection{Communication volume}
\label{sec:comm_vol}

For \texttt{Ring Attention}'s P2P communication strategy, the total volume of data being communicated between devices (in units of number of tensor elements) per iteration scales with $p$ and is given by:

\begin{equation}
V_{ring} = 2bt d \times p
\end{equation}
where $p$ is the number of devices. The first factor comes from counting the total number of communicated elements corresponding to $\{(\mathbf{k}_{\hat{a}},\mathbf{v}_{\hat{a}}), \hat{a}=1,\cdots,t\}$, i.e. 
\begin{equation}
    \textrm{numel}\left(\{(\mathbf{k}_{\hat{a}},\mathbf{v}_{\hat{a}}), \hat{a}=1,\cdots,t\} \right) = 2 btd.
\end{equation}
The \texttt{Allreduce} strategy we use in \texttt{Tree Decoding} requires the following volume \cite{Comms}: 

\begin{equation}
V_{\texttt{Allreduce}} = 2\times \frac{p-1}{p}\times \textrm{numel}.
\end{equation}
We communicate a shard of the numerator, denominator and max, requiring:

\begin{equation}
\textrm{numel}\left(\textrm{n},\textrm{d},m\right) = bd + 2bn_h.
\end{equation}
Note that we first perform on device the local reductions to obtain the local numerator and denominator on each device which consequently makes it so that $t$, i.e. the size of the local sequence chunk does not appear in the above expression. We then obtain: 

\begin{equation}
V_{Tree}= 2\frac{p-1}{p} \times\left(b d + 2bn_h\right).
\end{equation}

Our theoretical analysis shows that per iteration our algorithm maintains a lower communication volume than \texttt{Ring Attention}. Note however that \texttt{Ring Attention} when performed in the training setting with many queries overlaps communication and computation so as to hide its communication costs. 
However, overlapping communication and computation in the decoding case is infeasible because of how fast the attention computation on a single GPU is relative to how long it takes to communicate the chunk of keys and values between two devices. 

Concretely, let us take the example of decoding from a context of length 640000 split between $8$ GPUs within one node. Let us take a hidden size of 2048 and fix our data type to be \texttt{bfloat16}. Each device for decoding takes $O(10^{-5})$ seconds to perform  
the \texttt{Flash Attention} computation. 
The time it takes to move the keys and values of the corresponding size between adjacent GPUs as per Fig.~\ref{fig:nccl-pt2pt} is roughly $O(10^{-3})$ seconds. The latency incurred between nodes is even greater and therefore overlapping is not feasible due to this disparity in timescales.

\subsection{Performance with a LLama Transformer Model}
\label{sec:llama_perf}
To show that \texttt{Tree attention} can also be used in real world applications, we also measured end-to-end throughput with the Llama 3.1 8B model \cite{grattafiori2024llama3herdmodels} on prompt sequences of length 32k, 64k, 128k and 256k using ring attention or tree attention for decoding (with prefill) 10 tokens in Table~\ref{tab:decoding_times}. 
We ran these experiments on 8 H100 GPUs in a DGX cluster (connected with NVLink) as well as 4 MI300X GPUs in an AMD cluster connected with AMD infinity fabric. 
In Table~\ref{tab:decoding_times_4090} of Appendix~\ref{app:llama_perf}, we also show similar throughput results on 2  RTX 4090 GPUs connected with PCIe. In all cases we see that  \texttt{Tree attention} for decoding has significantly lower latency than  \texttt{Ring Attention} for decoding with a prefill stage.
\texttt{Ring Attention} is up to $\times$4 faster using 8x H100s and up to $\times$3 faster using 4x MI300x.
We expect this gap to increase as we increase the number of nodes.

While we have previously discussed that \texttt{Ring Attention} works best when used with the Ring Topology of TPU clusters, Table~\ref{tab:decoding_times} and \ref{tab:decoding_times_4090} show that \texttt{Tree Attention} results generalize well to various types of systems, number of GPUs, communication protocols and network topologies.

\begin{table}[ht]
\centering
\caption{Average Decoding Time  (in seconds) with a prefill stage comparisons, using the 8B Llama 3.1 model with \texttt{Tree Attention} (ours) and \texttt{Ring Attention} (SOTA) across various sequence lengths and GPU types. Average results and standard error ($\pm$) are computed using 10 trial runs.}
\begin{tabular}{@{}c@{\hskip 15pt}c@{\hskip 8pt}c@{\hskip 6pt}c@{\hskip 2pt}c@{\hskip 10pt}c@{\hskip 8pt}c@{\hskip 6pt}c@{\hskip 2pt}}
\toprule
\multirow{2}{*}{Sequence} & \multicolumn{3}{c}{\textbf{8x H100s}} & \phantom{abc} & \multicolumn{3}{c}{\textbf{4x MI300x}} \\ \cmidrule{2-4} \cmidrule{6-8}
 Length & \texttt{Tree Attn} & \texttt{Ring Attn} & Speedup & & \texttt{Tree Attn} & \texttt{Ring Attn} & Speedup \\ \midrule
32k  & \textbf{0.60} $\pm$ \small 0.15  & 2.57 $\pm$ \small 0.35 &  $\times$4 && \textbf{1.05} $\pm$ \small 0.01  & 3.57 $\pm$ \small 0.25  & $\times$3   \\
64k  & \textbf{1.08} $\pm$ \small 0.10  & 4.42 $\pm$ \small 0.38  & $\times$4 && \textbf{2.36} $\pm$ \small 0.01  & 7.33 $\pm$ \small 0.25  & $\times$3  \\
128k & \textbf{2.68} $\pm$ \small 0.28  & 6.38 $\pm$ \small 0.58  & $\times$2 && \textbf{6.43} $\pm$ \small 0.25  & 16.40 $\pm$ \small 0.40  & $\times$3  \\
256k & \textbf{2.89} $\pm$ \small 0.62  & 8.19 $\pm$ \small 1.07  & $\times$3 && \textbf{15.30} $\pm$ \small 4.93  & 35.12 $\pm$ \small 5.02  & $\times$2 \\
\bottomrule
\end{tabular}
\label{tab:decoding_times}
\end{table}

\section{Discussion and Conclusion}
\label{sec:discussion}

In this paper, we have derived the energy function for self-attention and demonstrated how the computation of the derivative of this function provides a novel and efficient method for computing attention in parallel. 
This advantage is especially apparent when performing decoding across multiple devices, in which case our \texttt{Tree Attention} enables us to substantially outperform SOTA \texttt{Ring Attention} with an \emph{asymptotically} superior algorithm, with $\times8$ speedups when we use 128 GPUs on a sequence length of 5.12M.
We also see that the \texttt{AllReduce} operation that we use involves sending partially reduced objects, which greatly reduces the volume of communicated data as well as the peak memory requirement.
In a real-world application, using the Llama 3.1 model with 1B and 8B parameters, we find that decoding with a prefill stage using \texttt{Tree Attention} gets us $\times$3-5 speedupds compared to \texttt{Ring Attention}.
Further, by testing our method on various types of GPUs clusters including AMD MI300xs, we show that \texttt{Tree Attention} generalizes very well to various communication protocols and network topologies.




\clearpage
\bibliography{iclr2025_conference}

\begin{thebibliography}{64}
\providecommand{\natexlab}[1]{#1}
\providecommand{\url}[1]{\texttt{#1}}
\expandafter\ifx\csname urlstyle\endcsname\relax
  \providecommand{\doi}[1]{doi: #1}\else
  \providecommand{\doi}{doi: \begingroup \urlstyle{rm}\Url}\fi

\bibitem[kai(2023)]{kaiokendev}
Things i’m learning while training superhot, 2023.
\newblock URL \url{https://kaiokendev.github.io/til}.

\bibitem[fla(2024)]{flashdecoding}
Flash-decoding for long-context inference, 2024.
\newblock URL \url{https://princeton-nlp.github.io/flash-decoding}.

\bibitem[Achiam et~al.(2023)Achiam, Adler, Agarwal, Ahmad, Akkaya, Aleman, Almeida, Altenschmidt, Altman, Anadkat, et~al.]{achiam2023gpt}
Josh Achiam, Steven Adler, Sandhini Agarwal, Lama Ahmad, Ilge Akkaya, Florencia~Leoni Aleman, Diogo Almeida, Janko Altenschmidt, Sam Altman, Shyamal Anadkat, et~al.
\newblock Gpt-4 technical report.
\newblock \emph{arXiv preprint arXiv:2303.08774}, 2023.

\bibitem[Anthony et~al.(2024)Anthony, Michalowicz, J.~Hatef, Abduljabbar, A.~Shafi, and Panda]{Comms}
Q.~Anthony, B.~Michalowicz, L.~Xu J.~Hatef, M.~Abduljabbar, H.~Subramoni A.~Shafi, and D.~Panda.
\newblock Demystifying the communication characteristics for distributed transformer models.
\newblock August 2024.

\bibitem[Arora et~al.(2024)Arora, Eyuboglu, Zhang, Timalsina, Alberti, Zinsley, Zou, Rudra, and R{\'e}]{arora2024simple}
Simran Arora, Sabri Eyuboglu, Michael Zhang, Aman Timalsina, Silas Alberti, Dylan Zinsley, James Zou, Atri Rudra, and Christopher R{\'e}.
\newblock Simple linear attention language models balance the recall-throughput tradeoff.
\newblock \emph{arXiv preprint arXiv:2402.18668}, 2024.

\bibitem[Bahdanau et~al.(2014)Bahdanau, Cho, and Bengio]{bahdanau2014neural}
Dzmitry Bahdanau, Kyunghyun Cho, and Yoshua Bengio.
\newblock Neural machine translation by jointly learning to align and translate.
\newblock \emph{arXiv preprint arXiv:1409.0473}, 2014.

\bibitem[Beal(2003)]{beal2003variational}
Matthew~James Beal.
\newblock \emph{Variational algorithms for approximate Bayesian inference}.
\newblock University of London, University College London (United Kingdom), 2003.

\bibitem[Beltagy et~al.(2020)Beltagy, Peters, and Cohan]{longformer}
Iz~Beltagy, Matthew~E. Peters, and Arman Cohan.
\newblock Longformer: The long-document transformer.
\newblock \emph{CoRR}, abs/2004.05150, 2020.
\newblock URL \url{https://arxiv.org/abs/2004.05150}.

\bibitem[Bertsch et~al.(2024)Bertsch, Ivgi, Alon, Berant, Gormley, and Neubig]{bertsch2024context}
Amanda Bertsch, Maor Ivgi, Uri Alon, Jonathan Berant, Matthew~R Gormley, and Graham Neubig.
\newblock In-context learning with long-context models: An in-depth exploration.
\newblock \emph{arXiv preprint arXiv:2405.00200}, 2024.

\bibitem[Betker(2023)]{betker2023better}
James Betker.
\newblock Better speech synthesis through scaling.
\newblock \emph{arXiv preprint arXiv:2305.07243}, 2023.

\bibitem[Brown et~al.(2020)Brown, Mann, Ryder, Subbiah, Kaplan, Dhariwal, Neelakantan, Shyam, Sastry, Askell, et~al.]{brown2020language}
Tom Brown, Benjamin Mann, Nick Ryder, Melanie Subbiah, Jared~D Kaplan, Prafulla Dhariwal, Arvind Neelakantan, Pranav Shyam, Girish Sastry, Amanda Askell, et~al.
\newblock Language models are few-shot learners.
\newblock \emph{Advances in neural information processing systems}, 33:\penalty0 1877--1901, 2020.

\bibitem[{Character AI}(2024)]{CharacterAI2024}
{Character AI}.
\newblock Optimizing inference, 2024.
\newblock URL \url{https://research.character.ai/optimizing-inference/}.

\bibitem[Chen et~al.(2021)Chen, Lu, Rajeswaran, Lee, Grover, Laskin, Abbeel, Srinivas, and Mordatch]{chen2021decision}
Lili Chen, Kevin Lu, Aravind Rajeswaran, Kimin Lee, Aditya Grover, Misha Laskin, Pieter Abbeel, Aravind Srinivas, and Igor Mordatch.
\newblock Decision transformer: Reinforcement learning via sequence modeling.
\newblock \emph{Advances in neural information processing systems}, 34:\penalty0 15084--15097, 2021.

\bibitem[Chen et~al.(2023)Chen, Wong, Chen, and Tian]{chen2023extending}
Shouyuan Chen, Sherman Wong, Liangjian Chen, and Yuandong Tian.
\newblock Extending context window of large language models via positional interpolation.
\newblock \emph{arXiv preprint arXiv:2306.15595}, 2023.

\bibitem[Choromanski et~al.(2020{\natexlab{a}})Choromanski, Likhosherstov, Dohan, Song, Gane, Sarl{\'{o}}s, Hawkins, Davis, Mohiuddin, Kaiser, Belanger, Colwell, and Weller]{performer}
Krzysztof Choromanski, Valerii Likhosherstov, David Dohan, Xingyou Song, Andreea Gane, Tam{\'{a}}s Sarl{\'{o}}s, Peter Hawkins, Jared Davis, Afroz Mohiuddin, Lukasz Kaiser, David Belanger, Lucy~J. Colwell, and Adrian Weller.
\newblock Rethinking attention with performers.
\newblock \emph{CoRR}, abs/2009.14794, 2020{\natexlab{a}}.
\newblock URL \url{https://arxiv.org/abs/2009.14794}.

\bibitem[Choromanski et~al.(2020{\natexlab{b}})Choromanski, Likhosherstov, Dohan, Song, Gane, Sarlos, Hawkins, Davis, Mohiuddin, Kaiser, et~al.]{choromanski2020rethinking}
Krzysztof Choromanski, Valerii Likhosherstov, David Dohan, Xingyou Song, Andreea Gane, Tamas Sarlos, Peter Hawkins, Jared Davis, Afroz Mohiuddin, Lukasz Kaiser, et~al.
\newblock Rethinking attention with performers.
\newblock \emph{arXiv preprint arXiv:2009.14794}, 2020{\natexlab{b}}.

\bibitem[D'Amico \& Negri(2024)D'Amico and Negri]{damico2024selfattentionattractornetworktransient}
Francesco D'Amico and Matteo Negri.
\newblock Self-attention as an attractor network: transient memories without backpropagation, 2024.
\newblock URL \url{https://arxiv.org/abs/2409.16112}.

\bibitem[Dao(2023)]{dao2023flashattention}
Tri Dao.
\newblock Flashattention-2: Faster attention with better parallelism and work partitioning.
\newblock \emph{arXiv preprint arXiv:2307.08691}, 2023.

\bibitem[Dao \& Gu(2024)Dao and Gu]{dao2024transformers}
Tri Dao and Albert Gu.
\newblock Transformers are ssms: Generalized models and efficient algorithms through structured state space duality.
\newblock \emph{arXiv preprint arXiv:2405.21060}, 2024.

\bibitem[Dao et~al.(2022)Dao, Fu, Ermon, Rudra, and R{\'e}]{dao2022flashattention}
Tri Dao, Dan Fu, Stefano Ermon, Atri Rudra, and Christopher R{\'e}.
\newblock Flashattention: Fast and memory-efficient exact attention with io-awareness.
\newblock \emph{Advances in Neural Information Processing Systems}, 35:\penalty0 16344--16359, 2022.

\bibitem[Dosovitskiy et~al.(2020)Dosovitskiy, Beyer, Kolesnikov, Weissenborn, Zhai, Unterthiner, Dehghani, Minderer, Heigold, Gelly, et~al.]{dosovitskiy2020image}
Alexey Dosovitskiy, Lucas Beyer, Alexander Kolesnikov, Dirk Weissenborn, Xiaohua Zhai, Thomas Unterthiner, Mostafa Dehghani, Matthias Minderer, Georg Heigold, Sylvain Gelly, et~al.
\newblock An image is worth 16x16 words: Transformers for image recognition at scale.
\newblock \emph{arXiv preprint arXiv:2010.11929}, 2020.

\bibitem[{Duman Keles} et~al.(2022){Duman Keles}, {Mahesakya Wijewardena}, and {Hegde}]{Keles}
Feyza {Duman Keles}, Pruthuvi {Mahesakya Wijewardena}, and Chinmay {Hegde}.
\newblock {On The Computational Complexity of Self-Attention}.
\newblock \emph{arXiv e-prints}, art. arXiv:2209.04881, September 2022.
\newblock \doi{10.48550/arXiv.2209.04881}.

\bibitem[{Feng} et~al.(2024){Feng}, {Tung}, {Hajimirsadeghi}, {Osama Ahmed}, {Bengio}, and {Mori}]{Aaren}
Leo {Feng}, Frederick {Tung}, Hossein {Hajimirsadeghi}, Mohamed {Osama Ahmed}, Yoshua {Bengio}, and Greg {Mori}.
\newblock {Attention as an RNN}.
\newblock \emph{arXiv e-prints}, art. arXiv:2405.13956, May 2024.
\newblock \doi{10.48550/arXiv.2405.13956}.

\bibitem[Glorioso et~al.(2024)Glorioso, Anthony, Tokpanov, Whittington, Pilault, Ibrahim, and Millidge]{glorioso2024zamba}
Paolo Glorioso, Quentin Anthony, Yury Tokpanov, James Whittington, Jonathan Pilault, Adam Ibrahim, and Beren Millidge.
\newblock Zamba: A compact 7b ssm hybrid model.
\newblock \emph{arXiv preprint arXiv:2405.16712}, 2024.

\bibitem[Grattafiori et~al.(2024)Grattafiori, Dubey, Jauhri, Pandey, Kadian, Al-Dahle, Letman, Mathur, Schelten, Vaughan, Yang, Fan, Goyal, Hartshorn, Yang, Mitra, Sravankumar, Korenev, Hinsvark, Rao, Zhang, Rodriguez, Gregerson, Spataru, Roziere, Biron, Tang, Chern, Caucheteux, Nayak, Bi, Marra, McConnell, Keller, Touret, Wu, Wong, Ferrer, Nikolaidis, Allonsius, Song, Pintz, Livshits, Wyatt, Esiobu, Choudhary, Mahajan, Garcia-Olano, Perino, Hupkes, Lakomkin, AlBadawy, Lobanova, Dinan, Smith, Radenovic, Guzmán, Zhang, Synnaeve, Lee, Anderson, Thattai, Nail, Mialon, Pang, Cucurell, Nguyen, Korevaar, Xu, Touvron, Zarov, Ibarra, Kloumann, Misra, Evtimov, Zhang, Copet, Lee, Geffert, Vranes, Park, Mahadeokar, Shah, van~der Linde, Billock, Hong, Lee, Fu, Chi, Huang, Liu, Wang, Yu, Bitton, Spisak, Park, Rocca, Johnstun, Saxe, Jia, Alwala, Prasad, Upasani, Plawiak, Li, Heafield, Stone, El-Arini, Iyer, Malik, Chiu, Bhalla, Lakhotia, Rantala-Yeary, van~der Maaten, Chen, Tan, Jenkins, Martin, Madaan, Malo, Blecher,
  Landzaat, de~Oliveira, Muzzi, Pasupuleti, Singh, Paluri, Kardas, Tsimpoukelli, Oldham, Rita, Pavlova, Kambadur, Lewis, Si, Singh, Hassan, Goyal, Torabi, Bashlykov, Bogoychev, Chatterji, Zhang, Duchenne, Çelebi, Alrassy, Zhang, Li, Vasic, Weng, Bhargava, Dubal, Krishnan, Koura, Xu, He, Dong, Srinivasan, Ganapathy, Calderer, Cabral, Stojnic, Raileanu, Maheswari, Girdhar, Patel, Sauvestre, Polidoro, Sumbaly, Taylor, Silva, Hou, Wang, Hosseini, Chennabasappa, Singh, Bell, Kim, Edunov, Nie, Narang, Raparthy, Shen, Wan, Bhosale, Zhang, Vandenhende, Batra, Whitman, Sootla, Collot, Gururangan, Borodinsky, Herman, Fowler, Sheasha, Georgiou, Scialom, Speckbacher, Mihaylov, Xiao, Karn, Goswami, Gupta, Ramanathan, Kerkez, Gonguet, Do, Vogeti, Albiero, Petrovic, Chu, Xiong, Fu, Meers, Martinet, Wang, Wang, Tan, Xia, Xie, Jia, Wang, Goldschlag, Gaur, Babaei, Wen, Song, Zhang, Li, Mao, Coudert, Yan, Chen, Papakipos, Singh, Srivastava, Jain, Kelsey, Shajnfeld, Gangidi, Victoria, Goldstand, Menon, Sharma, Boesenberg,
  Baevski, Feinstein, Kallet, Sangani, Teo, Yunus, Lupu, Alvarado, Caples, Gu, Ho, Poulton, Ryan, Ramchandani, Dong, Franco, Goyal, Saraf, Chowdhury, Gabriel, Bharambe, Eisenman, Yazdan, James, Maurer, Leonhardi, Huang, Loyd, Paola, Paranjape, Liu, Wu, Ni, Hancock, Wasti, Spence, Stojkovic, Gamido, Montalvo, Parker, Burton, Mejia, Liu, Wang, Kim, Zhou, Hu, Chu, Cai, Tindal, Feichtenhofer, Gao, Civin, Beaty, Kreymer, Li, Adkins, Xu, Testuggine, David, Parikh, Liskovich, Foss, Wang, Le, Holland, Dowling, Jamil, Montgomery, Presani, Hahn, Wood, Le, Brinkman, Arcaute, Dunbar, Smothers, Sun, Kreuk, Tian, Kokkinos, Ozgenel, Caggioni, Kanayet, Seide, Florez, Schwarz, Badeer, Swee, Halpern, Herman, Sizov, Guangyi, Zhang, Lakshminarayanan, Inan, Shojanazeri, Zou, Wang, Zha, Habeeb, Rudolph, Suk, Aspegren, Goldman, Zhan, Damlaj, Molybog, Tufanov, Leontiadis, Veliche, Gat, Weissman, Geboski, Kohli, Lam, Asher, Gaya, Marcus, Tang, Chan, Zhen, Reizenstein, Teboul, Zhong, Jin, Yang, Cummings, Carvill, Shepard, McPhie,
  Torres, Ginsburg, Wang, Wu, U, Saxena, Khandelwal, Zand, Matosich, Veeraraghavan, Michelena, Li, Jagadeesh, Huang, Chawla, Huang, Chen, Garg, A, Silva, Bell, Zhang, Guo, Yu, Moshkovich, Wehrstedt, Khabsa, Avalani, Bhatt, Mankus, Hasson, Lennie, Reso, Groshev, Naumov, Lathi, Keneally, Liu, Seltzer, Valko, Restrepo, Patel, Vyatskov, Samvelyan, Clark, Macey, Wang, Hermoso, Metanat, Rastegari, Bansal, Santhanam, Parks, White, Bawa, Singhal, Egebo, Usunier, Mehta, Laptev, Dong, Cheng, Chernoguz, Hart, Salpekar, Kalinli, Kent, Parekh, Saab, Balaji, Rittner, Bontrager, Roux, Dollar, Zvyagina, Ratanchandani, Yuvraj, Liang, Alao, Rodriguez, Ayub, Murthy, Nayani, Mitra, Parthasarathy, Li, Hogan, Battey, Wang, Howes, Rinott, Mehta, Siby, Bondu, Datta, Chugh, Hunt, Dhillon, Sidorov, Pan, Mahajan, Verma, Yamamoto, Ramaswamy, Lindsay, Lindsay, Feng, Lin, Zha, Patil, Shankar, Zhang, Zhang, Wang, Agarwal, Sajuyigbe, Chintala, Max, Chen, Kehoe, Satterfield, Govindaprasad, Gupta, Deng, Cho, Virk, Subramanian, Choudhury,
  Goldman, Remez, Glaser, Best, Koehler, Robinson, Li, Zhang, Matthews, Chou, Shaked, Vontimitta, Ajayi, Montanez, Mohan, Kumar, Mangla, Ionescu, Poenaru, Mihailescu, Ivanov, Li, Wang, Jiang, Bouaziz, Constable, Tang, Wu, Wang, Wu, Gao, Kleinman, Chen, Hu, Jia, Qi, Li, Zhang, Zhang, Adi, Nam, Yu, Wang, Zhao, Hao, Qian, Li, He, Rait, DeVito, Rosnbrick, Wen, Yang, Zhao, and Ma]{grattafiori2024llama3herdmodels}
Aaron Grattafiori, Abhimanyu Dubey, Abhinav Jauhri, Abhinav Pandey, Abhishek Kadian, Ahmad Al-Dahle, Aiesha Letman, Akhil Mathur, Alan Schelten, Alex Vaughan, Amy Yang, Angela Fan, Anirudh Goyal, Anthony Hartshorn, Aobo Yang, Archi Mitra, Archie Sravankumar, Artem Korenev, Arthur Hinsvark, Arun Rao, Aston Zhang, Aurelien Rodriguez, Austen Gregerson, Ava Spataru, Baptiste Roziere, Bethany Biron, Binh Tang, Bobbie Chern, Charlotte Caucheteux, Chaya Nayak, Chloe Bi, Chris Marra, Chris McConnell, Christian Keller, Christophe Touret, Chunyang Wu, Corinne Wong, Cristian~Canton Ferrer, Cyrus Nikolaidis, Damien Allonsius, Daniel Song, Danielle Pintz, Danny Livshits, Danny Wyatt, David Esiobu, Dhruv Choudhary, Dhruv Mahajan, Diego Garcia-Olano, Diego Perino, Dieuwke Hupkes, Egor Lakomkin, Ehab AlBadawy, Elina Lobanova, Emily Dinan, Eric~Michael Smith, Filip Radenovic, Francisco Guzmán, Frank Zhang, Gabriel Synnaeve, Gabrielle Lee, Georgia~Lewis Anderson, Govind Thattai, Graeme Nail, Gregoire Mialon, Guan Pang,
  Guillem Cucurell, Hailey Nguyen, Hannah Korevaar, Hu~Xu, Hugo Touvron, Iliyan Zarov, Imanol~Arrieta Ibarra, Isabel Kloumann, Ishan Misra, Ivan Evtimov, Jack Zhang, Jade Copet, Jaewon Lee, Jan Geffert, Jana Vranes, Jason Park, Jay Mahadeokar, Jeet Shah, Jelmer van~der Linde, Jennifer Billock, Jenny Hong, Jenya Lee, Jeremy Fu, Jianfeng Chi, Jianyu Huang, Jiawen Liu, Jie Wang, Jiecao Yu, Joanna Bitton, Joe Spisak, Jongsoo Park, Joseph Rocca, Joshua Johnstun, Joshua Saxe, Junteng Jia, Kalyan~Vasuden Alwala, Karthik Prasad, Kartikeya Upasani, Kate Plawiak, Ke~Li, Kenneth Heafield, Kevin Stone, Khalid El-Arini, Krithika Iyer, Kshitiz Malik, Kuenley Chiu, Kunal Bhalla, Kushal Lakhotia, Lauren Rantala-Yeary, Laurens van~der Maaten, Lawrence Chen, Liang Tan, Liz Jenkins, Louis Martin, Lovish Madaan, Lubo Malo, Lukas Blecher, Lukas Landzaat, Luke de~Oliveira, Madeline Muzzi, Mahesh Pasupuleti, Mannat Singh, Manohar Paluri, Marcin Kardas, Maria Tsimpoukelli, Mathew Oldham, Mathieu Rita, Maya Pavlova, Melanie Kambadur,
  Mike Lewis, Min Si, Mitesh~Kumar Singh, Mona Hassan, Naman Goyal, Narjes Torabi, Nikolay Bashlykov, Nikolay Bogoychev, Niladri Chatterji, Ning Zhang, Olivier Duchenne, Onur Çelebi, Patrick Alrassy, Pengchuan Zhang, Pengwei Li, Petar Vasic, Peter Weng, Prajjwal Bhargava, Pratik Dubal, Praveen Krishnan, Punit~Singh Koura, Puxin Xu, Qing He, Qingxiao Dong, Ragavan Srinivasan, Raj Ganapathy, Ramon Calderer, Ricardo~Silveira Cabral, Robert Stojnic, Roberta Raileanu, Rohan Maheswari, Rohit Girdhar, Rohit Patel, Romain Sauvestre, Ronnie Polidoro, Roshan Sumbaly, Ross Taylor, Ruan Silva, Rui Hou, Rui Wang, Saghar Hosseini, Sahana Chennabasappa, Sanjay Singh, Sean Bell, Seohyun~Sonia Kim, Sergey Edunov, Shaoliang Nie, Sharan Narang, Sharath Raparthy, Sheng Shen, Shengye Wan, Shruti Bhosale, Shun Zhang, Simon Vandenhende, Soumya Batra, Spencer Whitman, Sten Sootla, Stephane Collot, Suchin Gururangan, Sydney Borodinsky, Tamar Herman, Tara Fowler, Tarek Sheasha, Thomas Georgiou, Thomas Scialom, Tobias Speckbacher,
  Todor Mihaylov, Tong Xiao, Ujjwal Karn, Vedanuj Goswami, Vibhor Gupta, Vignesh Ramanathan, Viktor Kerkez, Vincent Gonguet, Virginie Do, Vish Vogeti, Vítor Albiero, Vladan Petrovic, Weiwei Chu, Wenhan Xiong, Wenyin Fu, Whitney Meers, Xavier Martinet, Xiaodong Wang, Xiaofang Wang, Xiaoqing~Ellen Tan, Xide Xia, Xinfeng Xie, Xuchao Jia, Xuewei Wang, Yaelle Goldschlag, Yashesh Gaur, Yasmine Babaei, Yi~Wen, Yiwen Song, Yuchen Zhang, Yue Li, Yuning Mao, Zacharie~Delpierre Coudert, Zheng Yan, Zhengxing Chen, Zoe Papakipos, Aaditya Singh, Aayushi Srivastava, Abha Jain, Adam Kelsey, Adam Shajnfeld, Adithya Gangidi, Adolfo Victoria, Ahuva Goldstand, Ajay Menon, Ajay Sharma, Alex Boesenberg, Alexei Baevski, Allie Feinstein, Amanda Kallet, Amit Sangani, Amos Teo, Anam Yunus, Andrei Lupu, Andres Alvarado, Andrew Caples, Andrew Gu, Andrew Ho, Andrew Poulton, Andrew Ryan, Ankit Ramchandani, Annie Dong, Annie Franco, Anuj Goyal, Aparajita Saraf, Arkabandhu Chowdhury, Ashley Gabriel, Ashwin Bharambe, Assaf Eisenman, Azadeh
  Yazdan, Beau James, Ben Maurer, Benjamin Leonhardi, Bernie Huang, Beth Loyd, Beto~De Paola, Bhargavi Paranjape, Bing Liu, Bo~Wu, Boyu Ni, Braden Hancock, Bram Wasti, Brandon Spence, Brani Stojkovic, Brian Gamido, Britt Montalvo, Carl Parker, Carly Burton, Catalina Mejia, Ce~Liu, Changhan Wang, Changkyu Kim, Chao Zhou, Chester Hu, Ching-Hsiang Chu, Chris Cai, Chris Tindal, Christoph Feichtenhofer, Cynthia Gao, Damon Civin, Dana Beaty, Daniel Kreymer, Daniel Li, David Adkins, David Xu, Davide Testuggine, Delia David, Devi Parikh, Diana Liskovich, Didem Foss, Dingkang Wang, Duc Le, Dustin Holland, Edward Dowling, Eissa Jamil, Elaine Montgomery, Eleonora Presani, Emily Hahn, Emily Wood, Eric-Tuan Le, Erik Brinkman, Esteban Arcaute, Evan Dunbar, Evan Smothers, Fei Sun, Felix Kreuk, Feng Tian, Filippos Kokkinos, Firat Ozgenel, Francesco Caggioni, Frank Kanayet, Frank Seide, Gabriela~Medina Florez, Gabriella Schwarz, Gada Badeer, Georgia Swee, Gil Halpern, Grant Herman, Grigory Sizov, Guangyi, Zhang, Guna
  Lakshminarayanan, Hakan Inan, Hamid Shojanazeri, Han Zou, Hannah Wang, Hanwen Zha, Haroun Habeeb, Harrison Rudolph, Helen Suk, Henry Aspegren, Hunter Goldman, Hongyuan Zhan, Ibrahim Damlaj, Igor Molybog, Igor Tufanov, Ilias Leontiadis, Irina-Elena Veliche, Itai Gat, Jake Weissman, James Geboski, James Kohli, Janice Lam, Japhet Asher, Jean-Baptiste Gaya, Jeff Marcus, Jeff Tang, Jennifer Chan, Jenny Zhen, Jeremy Reizenstein, Jeremy Teboul, Jessica Zhong, Jian Jin, Jingyi Yang, Joe Cummings, Jon Carvill, Jon Shepard, Jonathan McPhie, Jonathan Torres, Josh Ginsburg, Junjie Wang, Kai Wu, Kam~Hou U, Karan Saxena, Kartikay Khandelwal, Katayoun Zand, Kathy Matosich, Kaushik Veeraraghavan, Kelly Michelena, Keqian Li, Kiran Jagadeesh, Kun Huang, Kunal Chawla, Kyle Huang, Lailin Chen, Lakshya Garg, Lavender A, Leandro Silva, Lee Bell, Lei Zhang, Liangpeng Guo, Licheng Yu, Liron Moshkovich, Luca Wehrstedt, Madian Khabsa, Manav Avalani, Manish Bhatt, Martynas Mankus, Matan Hasson, Matthew Lennie, Matthias Reso, Maxim
  Groshev, Maxim Naumov, Maya Lathi, Meghan Keneally, Miao Liu, Michael~L. Seltzer, Michal Valko, Michelle Restrepo, Mihir Patel, Mik Vyatskov, Mikayel Samvelyan, Mike Clark, Mike Macey, Mike Wang, Miquel~Jubert Hermoso, Mo~Metanat, Mohammad Rastegari, Munish Bansal, Nandhini Santhanam, Natascha Parks, Natasha White, Navyata Bawa, Nayan Singhal, Nick Egebo, Nicolas Usunier, Nikhil Mehta, Nikolay~Pavlovich Laptev, Ning Dong, Norman Cheng, Oleg Chernoguz, Olivia Hart, Omkar Salpekar, Ozlem Kalinli, Parkin Kent, Parth Parekh, Paul Saab, Pavan Balaji, Pedro Rittner, Philip Bontrager, Pierre Roux, Piotr Dollar, Polina Zvyagina, Prashant Ratanchandani, Pritish Yuvraj, Qian Liang, Rachad Alao, Rachel Rodriguez, Rafi Ayub, Raghotham Murthy, Raghu Nayani, Rahul Mitra, Rangaprabhu Parthasarathy, Raymond Li, Rebekkah Hogan, Robin Battey, Rocky Wang, Russ Howes, Ruty Rinott, Sachin Mehta, Sachin Siby, Sai~Jayesh Bondu, Samyak Datta, Sara Chugh, Sara Hunt, Sargun Dhillon, Sasha Sidorov, Satadru Pan, Saurabh Mahajan,
  Saurabh Verma, Seiji Yamamoto, Sharadh Ramaswamy, Shaun Lindsay, Shaun Lindsay, Sheng Feng, Shenghao Lin, Shengxin~Cindy Zha, Shishir Patil, Shiva Shankar, Shuqiang Zhang, Shuqiang Zhang, Sinong Wang, Sneha Agarwal, Soji Sajuyigbe, Soumith Chintala, Stephanie Max, Stephen Chen, Steve Kehoe, Steve Satterfield, Sudarshan Govindaprasad, Sumit Gupta, Summer Deng, Sungmin Cho, Sunny Virk, Suraj Subramanian, Sy~Choudhury, Sydney Goldman, Tal Remez, Tamar Glaser, Tamara Best, Thilo Koehler, Thomas Robinson, Tianhe Li, Tianjun Zhang, Tim Matthews, Timothy Chou, Tzook Shaked, Varun Vontimitta, Victoria Ajayi, Victoria Montanez, Vijai Mohan, Vinay~Satish Kumar, Vishal Mangla, Vlad Ionescu, Vlad Poenaru, Vlad~Tiberiu Mihailescu, Vladimir Ivanov, Wei Li, Wenchen Wang, Wenwen Jiang, Wes Bouaziz, Will Constable, Xiaocheng Tang, Xiaojian Wu, Xiaolan Wang, Xilun Wu, Xinbo Gao, Yaniv Kleinman, Yanjun Chen, Ye~Hu, Ye~Jia, Ye~Qi, Yenda Li, Yilin Zhang, Ying Zhang, Yossi Adi, Youngjin Nam, Yu, Wang, Yu~Zhao, Yuchen Hao, Yundi
  Qian, Yunlu Li, Yuzi He, Zach Rait, Zachary DeVito, Zef Rosnbrick, Zhaoduo Wen, Zhenyu Yang, Zhiwei Zhao, and Zhiyu Ma.
\newblock The llama 3 herd of models, 2024.
\newblock URL \url{https://arxiv.org/abs/2407.21783}.

\bibitem[Gu \& Dao(2023)Gu and Dao]{gu2023mamba}
Albert Gu and Tri Dao.
\newblock Mamba: Linear-time sequence modeling with selective state spaces.
\newblock \emph{arXiv preprint arXiv:2312.00752}, 2023.

\bibitem[Hoffmann et~al.(2022)Hoffmann, Borgeaud, Mensch, Buchatskaya, Cai, Rutherford, Casas, Hendricks, Welbl, Clark, et~al.]{hoffmann2022training}
Jordan Hoffmann, Sebastian Borgeaud, Arthur Mensch, Elena Buchatskaya, Trevor Cai, Eliza Rutherford, Diego de~Las Casas, Lisa~Anne Hendricks, Johannes Welbl, Aidan Clark, et~al.
\newblock Training compute-optimal large language models.
\newblock \emph{arXiv preprint arXiv:2203.15556}, 2022.

\bibitem[{Hoover} et~al.(2023){Hoover}, {Liang}, {Pham}, {Panda}, {Strobelt}, {Chau}, {Zaki}, and {Krotov}]{energy_transformer}
Benjamin {Hoover}, Yuchen {Liang}, Bao {Pham}, Rameswar {Panda}, Hendrik {Strobelt}, Duen~Horng {Chau}, Mohammed~J. {Zaki}, and Dmitry {Krotov}.
\newblock {Energy Transformer}.
\newblock \emph{arXiv e-prints}, art. arXiv:2302.07253, February 2023.
\newblock \doi{10.48550/arXiv.2302.07253}.

\bibitem[Hoover et~al.(2024)Hoover, Chau, Strobelt, Ram, and Krotov]{hoover2024dense}
Benjamin Hoover, Duen~Horng Chau, Hendrik Strobelt, Parikshit Ram, and Dmitry Krotov.
\newblock Dense associative memory through the lens of random features.
\newblock In \emph{The Thirty-eighth Annual Conference on Neural Information Processing Systems}, 2024.
\newblock URL \url{https://openreview.net/forum?id=164QnJsYjF}.

\bibitem[Hopfield(1982)]{energy_physical_sys_1982}
J~J Hopfield.
\newblock Neural networks and physical systems with emergent collective computational abilities.
\newblock \emph{Proceedings of the National Academy of Sciences}, 79\penalty0 (8):\penalty0 2554--2558, 1982.
\newblock \doi{10.1073/pnas.79.8.2554}.
\newblock URL \url{https://www.pnas.org/doi/abs/10.1073/pnas.79.8.2554}.

\bibitem[Kang et~al.(2024)Kang, Zhang, Kundu, Jeong, Liu, Krishna, and Zhao]{kang2024gearefficientkvcache}
Hao Kang, Qingru Zhang, Souvik Kundu, Geonhwa Jeong, Zaoxing Liu, Tushar Krishna, and Tuo Zhao.
\newblock Gear: An efficient kv cache compression recipe for near-lossless generative inference of llm, 2024.
\newblock URL \url{https://arxiv.org/abs/2403.05527}.

\bibitem[Kaplan et~al.(2020)Kaplan, McCandlish, Henighan, Brown, Chess, Child, Gray, Radford, Wu, and Amodei]{kaplan2020scaling}
Jared Kaplan, Sam McCandlish, Tom Henighan, Tom~B Brown, Benjamin Chess, Rewon Child, Scott Gray, Alec Radford, Jeffrey Wu, and Dario Amodei.
\newblock Scaling laws for neural language models.
\newblock \emph{arXiv preprint arXiv:2001.08361}, 2020.

\bibitem[Katharopoulos et~al.(2020)Katharopoulos, Vyas, Pappas, and Fleuret]{katharopoulos2020transformers}
Angelos Katharopoulos, Apoorv Vyas, Nikolaos Pappas, and Fran{\c{c}}ois Fleuret.
\newblock Transformers are rnns: Fast autoregressive transformers with linear attention.
\newblock In \emph{International conference on machine learning}, pp.\  5156--5165. PMLR, 2020.

\bibitem[Katsch(2023)]{katsch2023gateloop}
Tobias Katsch.
\newblock Gateloop: Fully data-controlled linear recurrence for sequence modeling.
\newblock \emph{arXiv preprint arXiv:2311.01927}, 2023.

\bibitem[{Krotov}(2021)]{HAM}
Dmitry {Krotov}.
\newblock {Hierarchical Associative Memory}.
\newblock \emph{arXiv e-prints}, art. arXiv:2107.06446, July 2021.
\newblock \doi{10.48550/arXiv.2107.06446}.

\bibitem[Krotov \& Hopfield(2016)Krotov and Hopfield]{krotov2016dense}
Dmitry Krotov and John~J Hopfield.
\newblock Dense associative memory for pattern recognition.
\newblock \emph{Advances in neural information processing systems}, 29, 2016.

\bibitem[Landau \& Lifshitz(1958)Landau and Lifshitz]{landau}
L.~D. Landau and E.~M. Lifshitz.
\newblock \emph{Statistical Physics}.
\newblock Reading, MA: Addison-Wesley, 1958.

\bibitem[LeCun et~al.(2006)LeCun, Chopra, Hadsell, Ranzato, Huang, et~al.]{lecun2006tutorial}
Yann LeCun, Sumit Chopra, Raia Hadsell, M~Ranzato, Fujie Huang, et~al.
\newblock A tutorial on energy-based learning.
\newblock \emph{Predicting structured data}, 1\penalty0 (0), 2006.

\bibitem[Lee et~al.(2024)Lee, Chen, Dai, Dua, Sachan, Boratko, Luan, Arnold, Perot, Dalmia, et~al.]{lee2024can}
Jinhyuk Lee, Anthony Chen, Zhuyun Dai, Dheeru Dua, Devendra~Singh Sachan, Michael Boratko, Yi~Luan, S{\'e}bastien~MR Arnold, Vincent Perot, Siddharth Dalmia, et~al.
\newblock Can long-context language models subsume retrieval, rag, sql, and more?
\newblock \emph{arXiv preprint arXiv:2406.13121}, 2024.

\bibitem[Liu et~al.(2024)Liu, Liu, Pan, He, Haffari, and Zhuang]{liu2024minicachekvcachecompression}
Akide Liu, Jing Liu, Zizheng Pan, Yefei He, Gholamreza Haffari, and Bohan Zhuang.
\newblock Minicache: Kv cache compression in depth dimension for large language models, 2024.
\newblock URL \url{https://arxiv.org/abs/2405.14366}.

\bibitem[{Liu} et~al.(2023){Liu}, {Zaharia}, and {Abbeel}]{Ring_Attn}
Hao {Liu}, Matei {Zaharia}, and Pieter {Abbeel}.
\newblock {Ring Attention with Blockwise Transformers for Near-Infinite Context}.
\newblock \emph{arXiv e-prints}, art. arXiv:2310.01889, October 2023.
\newblock \doi{10.48550/arXiv.2310.01889}.

\bibitem[Millidge et~al.(2022)Millidge, Salvatori, Song, Lukasiewicz, and Bogacz]{millidge2022universal}
Beren Millidge, Tommaso Salvatori, Yuhang Song, Thomas Lukasiewicz, and Rafal Bogacz.
\newblock Universal hopfield networks: A general framework for single-shot associative memory models.
\newblock In \emph{International Conference on Machine Learning}, pp.\  15561--15583. PMLR, 2022.

\bibitem[Nawrot et~al.(2024)Nawrot, Łańcucki, Chochowski, Tarjan, and Ponti]{nawrot2024dynamicmemorycompressionretrofitting}
Piotr Nawrot, Adrian Łańcucki, Marcin Chochowski, David Tarjan, and Edoardo~M. Ponti.
\newblock Dynamic memory compression: Retrofitting llms for accelerated inference, 2024.
\newblock URL \url{https://arxiv.org/abs/2403.09636}.

\bibitem[Peng et~al.(2023)Peng, Quesnelle, Fan, and Shippole]{peng2023yarn}
Bowen Peng, Jeffrey Quesnelle, Honglu Fan, and Enrico Shippole.
\newblock Yarn: Efficient context window extension of large language models.
\newblock \emph{arXiv preprint arXiv:2309.00071}, 2023.

\bibitem[Peng et~al.(2021)Peng, Pappas, Yogatama, Schwartz, Smith, and Kong]{peng2021random}
Hao Peng, Nikolaos Pappas, Dani Yogatama, Roy Schwartz, Noah~A Smith, and Lingpeng Kong.
\newblock Random feature attention.
\newblock \emph{arXiv preprint arXiv:2103.02143}, 2021.

\bibitem[Pilault et~al.(2023)Pilault, Liu, Bansal, and Dreyer]{ijcai2023p460}
Jonathan Pilault, Can Liu, Mohit Bansal, and Markus Dreyer.
\newblock On conditional and compositional language model differentiable prompting.
\newblock In Edith Elkind (ed.), \emph{Proceedings of the Thirty-Second International Joint Conference on Artificial Intelligence, {IJCAI-23}}, pp.\  4136--4144. International Joint Conferences on Artificial Intelligence Organization, 8 2023.
\newblock \doi{10.24963/ijcai.2023/460}.
\newblock URL \url{https://doi.org/10.24963/ijcai.2023/460}.
\newblock Main Track.

\bibitem[{Rabe} \& {Staats}(2021){Rabe} and {Staats}]{mem_attn}
Markus~N. {Rabe} and Charles {Staats}.
\newblock {Self-attention Does Not Need $O(n^2)$ Memory}.
\newblock \emph{arXiv e-prints}, art. arXiv:2112.05682, December 2021.
\newblock \doi{10.48550/arXiv.2112.05682}.

\bibitem[{Ramsauer} et~al.(2020){Ramsauer}, {Sch{\"a}fl}, {Lehner}, {Seidl}, {Widrich}, {Adler}, {Gruber}, {Holzleitner}, {Pavlovi{\'c}}, {Kjetil Sandve}, {Greiff}, {Kreil}, {Kopp}, {Klambauer}, {Brandstetter}, and {Hochreiter}]{Hopfield}
Hubert {Ramsauer}, Bernhard {Sch{\"a}fl}, Johannes {Lehner}, Philipp {Seidl}, Michael {Widrich}, Thomas {Adler}, Lukas {Gruber}, Markus {Holzleitner}, Milena {Pavlovi{\'c}}, Geir {Kjetil Sandve}, Victor {Greiff}, David {Kreil}, Michael {Kopp}, G{\"u}nter {Klambauer}, Johannes {Brandstetter}, and Sepp {Hochreiter}.
\newblock {Hopfield Networks is All You Need}.
\newblock \emph{arXiv e-prints}, art. arXiv:2008.02217, July 2020.
\newblock \doi{10.48550/arXiv.2008.02217}.

\bibitem[Reed et~al.(2022)Reed, Zolna, Parisotto, Colmenarejo, Novikov, Barth-Maron, Gimenez, Sulsky, Kay, Springenberg, et~al.]{reed2022generalist}
Scott Reed, Konrad Zolna, Emilio Parisotto, Sergio~Gomez Colmenarejo, Alexander Novikov, Gabriel Barth-Maron, Mai Gimenez, Yury Sulsky, Jackie Kay, Jost~Tobias Springenberg, et~al.
\newblock A generalist agent.
\newblock \emph{arXiv preprint arXiv:2205.06175}, 2022.

\bibitem[Reid et~al.(2024)Reid, Savinov, Teplyashin, Lepikhin, Lillicrap, Alayrac, Soricut, Lazaridou, Firat, Schrittwieser, et~al.]{reid2024gemini}
Machel Reid, Nikolay Savinov, Denis Teplyashin, Dmitry Lepikhin, Timothy Lillicrap, Jean-baptiste Alayrac, Radu Soricut, Angeliki Lazaridou, Orhan Firat, Julian Schrittwieser, et~al.
\newblock Gemini 1.5: Unlocking multimodal understanding across millions of tokens of context.
\newblock \emph{arXiv preprint arXiv:2403.05530}, 2024.

\bibitem[Shen et~al.(2021)Shen, Zhang, Zhao, Yi, and Li]{shen2021efficient}
Zhuoran Shen, Mingyuan Zhang, Haiyu Zhao, Shuai Yi, and Hongsheng Li.
\newblock Efficient attention: Attention with linear complexities.
\newblock In \emph{WACV}, 2021.

\bibitem[Singh \& Buckley(2023)Singh and Buckley]{singh2023attention}
Ryan Singh and Christopher~L Buckley.
\newblock Attention: Marginal probability is all you need?
\newblock \emph{arXiv preprint arXiv:2304.04556}, 2023.

\bibitem[Song \& Kingma(2021)Song and Kingma]{song2021train}
Yang Song and Diederik~P Kingma.
\newblock How to train your energy-based models.
\newblock \emph{arXiv preprint arXiv:2101.03288}, 2021.

\bibitem[Sun et~al.(2023)Sun, Dong, Huang, Ma, Xia, Xue, Wang, and Wei]{sun2023retentive}
Yutao Sun, Li~Dong, Shaohan Huang, Shuming Ma, Yuqing Xia, Jilong Xue, Jianyong Wang, and Furu Wei.
\newblock Retentive network: A successor to transformer for large language models.
\newblock \emph{arXiv preprint arXiv:2307.08621}, 2023.

\bibitem[Tay et~al.(2020)Tay, Dehghani, Bahri, and Metzler]{DBLP:journals/corr/abs-2009-06732}
Yi~Tay, Mostafa Dehghani, Dara Bahri, and Donald Metzler.
\newblock Efficient transformers: {A} survey.
\newblock \emph{CoRR}, abs/2009.06732, 2020.
\newblock URL \url{https://arxiv.org/abs/2009.06732}.

\bibitem[Team(2024)]{team2024chameleon}
Chameleon Team.
\newblock Chameleon: Mixed-modal early-fusion foundation models.
\newblock \emph{arXiv preprint arXiv:2405.09818}, 2024.

\bibitem[Team et~al.(2023)Team, Anil, Borgeaud, Wu, Alayrac, Yu, Soricut, Schalkwyk, Dai, Hauth, et~al.]{team2023gemini}
Gemini Team, Rohan Anil, Sebastian Borgeaud, Yonghui Wu, Jean-Baptiste Alayrac, Jiahui Yu, Radu Soricut, Johan Schalkwyk, Andrew~M Dai, Anja Hauth, et~al.
\newblock Gemini: a family of highly capable multimodal models.
\newblock \emph{arXiv preprint arXiv:2312.11805}, 2023.

\bibitem[Tsai et~al.(2019)Tsai, Bai, Yamada, Morency, and Salakhutdinov]{DBLP:journals/corr/abs-1908-11775}
Yao{-}Hung~Hubert Tsai, Shaojie Bai, Makoto Yamada, Louis{-}Philippe Morency, and Ruslan Salakhutdinov.
\newblock Transformer dissection: An unified understanding for transformer's attention via the lens of kernel.
\newblock \emph{CoRR}, abs/1908.11775, 2019.
\newblock URL \url{http://arxiv.org/abs/1908.11775}.

\bibitem[Vaswani et~al.(2017)Vaswani, Shazeer, Parmar, Uszkoreit, Jones, Gomez, Kaiser, and Polosukhin]{vaswani2017attention}
Ashish Vaswani, Noam Shazeer, Niki Parmar, Jakob Uszkoreit, Llion Jones, Aidan~N Gomez, {\L}ukasz Kaiser, and Illia Polosukhin.
\newblock Attention is all you need.
\newblock \emph{Advances in neural information processing systems}, 30, 2017.

\bibitem[Vieira(2016)]{f_v_gf}
Tim Vieira.
\newblock {Evaluating $\nabla f(x)$ is as fast as $f(x)$}, 2016.
\newblock \url{https://timvieira.github.io/blog/post/2016/09/25/evaluating-fx-is-as-fast-as-fx/}.

\bibitem[Wang et~al.(2020)Wang, Li, Khabsa, Fang, and Ma]{linformer}
Sinong Wang, Belinda~Z. Li, Madian Khabsa, Han Fang, and Hao Ma.
\newblock Linformer: Self-attention with linear complexity.
\newblock \emph{CoRR}, abs/2006.04768, 2020.
\newblock URL \url{https://arxiv.org/abs/2006.04768}.

\bibitem[Wu et~al.(2024)Wu, Tan, Wang, Wang, Li, and Sun]{wu2024beyond}
Shangda Wu, Xu~Tan, Zili Wang, Rui Wang, Xiaobing Li, and Maosong Sun.
\newblock Beyond language models: Byte models are digital world simulators.
\newblock \emph{arXiv preprint arXiv:2402.19155}, 2024.

\bibitem[Xue et~al.(2022)Xue, Barua, Constant, Al-Rfou, Narang, Kale, Roberts, and Raffel]{xue2022byt5}
Linting Xue, Aditya Barua, Noah Constant, Rami Al-Rfou, Sharan Narang, Mihir Kale, Adam Roberts, and Colin Raffel.
\newblock Byt5: Towards a token-free future with pre-trained byte-to-byte models.
\newblock \emph{Transactions of the Association for Computational Linguistics}, 10:\penalty0 291--306, 2022.

\bibitem[Zaheer et~al.(2020)Zaheer, Guruganesh, Dubey, Ainslie, Alberti, Onta{\~{n}}{\'{o}}n, Pham, Ravula, Wang, Yang, and Ahmed]{bigbird}
Manzil Zaheer, Guru Guruganesh, Avinava Dubey, Joshua Ainslie, Chris Alberti, Santiago Onta{\~{n}}{\'{o}}n, Philip Pham, Anirudh Ravula, Qifan Wang, Li~Yang, and Amr Ahmed.
\newblock Big bird: Transformers for longer sequences.
\newblock \emph{CoRR}, abs/2007.14062, 2020.
\newblock URL \url{https://arxiv.org/abs/2007.14062}.

\end{thebibliography}
\bibliographystyle{iclr2025_conference}

\clearpage
\appendix

\section{More Related Work}

Recent work has attempted a Bayesian reformulation of attention by deriving a probabilistic generative model which matches the operations performed in a self-attention operation \cite{singh2023attention}. 
This follows on from a line of work which relates self-attention to the well-studied Hopfield Network architecture \cite{Hopfield}. 
The idea being that while the Hopfield Network is defined in terms of dynamics upon an energy landscape, this same picture can be cast into a Bayesian interpretation by identifying the energy function with a variational free energy functional and hence deriving the generative model that is implicit in the self-attention update.

In particular, consider the energy function proposed in \cite{energy_transformer}, which is the logsumexp. 
Since the gradient in the update rule of that paper is taken with respect to the input to the block, the resulting function is a modified version of the self-attention operation. 
Similarly, the update rule in \cite{Hopfield} requires the tying of certain weights (K and V) in the attention operation. 
This restricts the Hopfield derivation to modelling auto-associative retrieval networks, while transformer attention is hetero-associative.

Another notable related work is \cite{Aaren}, where the authors made similar observations as we do in section \ref{sec:TreeAttn} about how the associative operations within the attention computation can be efficiently parallelized to motivate an attention-based modified RNN architecture for sequence modeling.

While this energy function by itself is primarily a mathematical and theoretical curiosity, we demonstrate below that when combined with automatic differentiation, our formulation naturally leads to highly efficient parallel algorithms for computing attention and performing decoding, especially across multiple devices.

\section{More background on the tree reduction operation}

A tree reduction operation is a hierarchical strategy to perform a reduction operation (e.g., sum, product, maximum, minimum) over a set of data elements efficiently, especially in parallel computing. This approach reduces the overall computational complexity and enables efficient utilization of parallel processing resources. Here's how it works:

\begin{itemize}
    \item Divide the problem into smaller tasks: The input data is divided into smaller chunks, and the reduction operation is performed pairwise between adjacent elements in these chunks.
    \item Form a tree-like structure: The results from the first level of reductions are themselves reduced pairwise in the next level. This continues until the entire dataset has been reduced to a single result.
    \item Iterative or recursive aggregation: The aggregation typically follows a binary tree pattern, but other fan-in numbers (e.g., k-ary trees) can also be used. Each node in the tree represents a partial reduction result, and the root of the tree holds the final result.
\end{itemize}

Because a tree structure has a logarithmic depth to total number of nodes, a tree reduction can asymptotiacally reduce the number of total steps required to perform an operation when it is possible to aggregate partial results, and additionally is amenable to parallelization since k-ary trees can be defined to match the number of available processors for parallel processing. Additionally, many existing networking topologies such as Nvidia's NVLINK and Infiniband, due to the natural advantages of tree structures, are designed with such a toplogy meaning that tree operations are natural and efficient to perform. 

\section{Attention as the Gradient of an Energy Function and Bayesian interpretations}

\subsection{Proof of Observation~\ref{obs:1}}
\label{app:proof_attn_energy}
Here, we show how the self-attention operation can be written as the gradient of an energy function. 
In particular, we define a scalar function that depends on the keys, queries, values and additionally on an auxiliary vector that we refer to as the \textit{source} $\zeta$. 
The source is the parameter with respect to which we compute the gradient of the scalar function to obtain the self-attention operation.
We need the source in order to write down the generating function of the moments of the distribution above. 
It is also the variable with respect to which we can Taylor-expand the generating function and extract the moments as the coefficients of the monomials of $\zeta$ appearing in the Taylor series.
Explicitly, we want to find a function $F(q,k,v,\zeta)$ such that:

\begin{equation}
\sum^N_{a=1}\softmax(q\cdot k_a)v_a = \frac{\partial F}{\partial \zeta}\bigg\vert_{\zeta=0}.
\end{equation}

This terminology is inspired by work on energy-based models in machine learning \cite{beal2003variational,lecun2006tutorial,song2021train}.
A summary of variables and indices is provided in appendix ~\ref{app:notations}

We first show how the energy function is given by the cumulant-generating function associated to the distribution given by attention scores. 
Taking inspiration from statistical mechanics, where an analogous cumulant-generating function defines the Helmholtz Free energy \citep{landau}, we dub our cumulant-generating function the \textit{energy function for self-attention}.

Let us focus on the case with a single query. 
As noted above, we leverage the fact that the attention operation can be seen as the computation of the expectation value of the vectors $v$ in the distribution set by the attention scores $z$:
\begin{equation}
z =\langle v\rangle = \sum^N_{a=1} P_a v_a=\frac{\sum^N_{a=1}e^{q\cdot k^T_a}v_{a}}{\sum^N_{i=1}e^{q\cdot k^T_i}}.
\end{equation}
The probability density is given by: 
\begin{equation}
P_a = \frac{e^{q\cdot k^T_a}}{\sum^N_{i=1}e^{q\cdot k^T_i}}.
\end{equation}
Typically, the denominator or normalization factor is identified with the so-called partition function: 
\begin{equation}
Z = \sum^N_{a=1}e^{q\cdot k^T_a}.
\end{equation}
We can now compute the first moment of the probability distribution given above by introducing a source, $\zeta\in \mathbb{R}^d$. 
In our case, with $\zeta,$ we can extend the partition function to the function: 
\begin{equation}
Z(\zeta)  = \sum^N_{a=1}e^{q\cdot k^T_a+\zeta\cdot v^T_a}.
\end{equation}
Now, we can compute any moment of the distribution as the $n$-th Taylor coefficient of $Z(\zeta)$ $\forall A_1,A_2,\cdots\in \{1,\cdots, d_h\}$ : 
\begin{equation}
\langle v_{A_1}\cdots v_{A_n}\rangle  = \frac{1}{Z}\frac{\partial^n Z(\zeta)}{\partial \zeta_{A_1}\cdots \partial \zeta_{A_n}}\bigg\vert_{\zeta=0}.
\end{equation}
In other words, we can write $Z(\zeta)$
as:
\begin{equation}
Z(\zeta) =Z\left( 1 + \langle v \rangle \zeta + \frac{1}{2!}\langle v_{A_1} v_{A_2}\rangle \zeta_{A_1}\zeta_{A_2}+ \cdots\right)
\end{equation}
 
Therefore, the first moment can be written as:
\begin{equation}
\langle v\rangle= \frac{1}{Z}\frac{\partial Z}{\partial \zeta}\bigg\vert_{\zeta=0},
\end{equation}
which can be written as the gradient of the log of $Z(\zeta)$:
\begin{equation}
\langle v\rangle = \frac{\partial}{\partial \zeta}\log Z(\zeta)\bigg\vert_{\zeta=0}.
\end{equation}
This quantity is the generating function, a.k.a. the free energy: 

\begin{equation}
F = \log \sum_a\exp\left(q\cdot k_a^T +\zeta\cdot v^T_a\right).
\end{equation}

To compute causal self-attention, we introduce $N$ sources $\zeta^i$ each $\in\mathbb{R}^{d}$ and take
\begin{equation}
F_{tot} = \sum^N_{i=1} F_{i} = \sum^N_{i=1}\log\sum^i_{a=1}\exp(q_i\cdot k^T_a+\zeta_i\cdot v^T_a).
\end{equation}
The truncation of the inner sum up to index $i$ is due to causal masking. 

Now, in order to compute the $i$-th element of causal self-attention, we differentiate with respect to $\zeta_i$ and set it to zero:
\begin{equation}
\frac{\partial F_{tot}}{\partial \zeta_{i,A}}\bigg\vert_{\zeta_i=0,\forall i} = \frac{\sum^i_{a=1}\exp(q^i\cdot k^T_a)v_{a,A}}{\sum^i_{a=1}\exp(q^i\cdot k^T_a)}.
\end{equation}
The generalization to the multi-head attention case is straightforward.
In this case, there is one key, query and value per head.
For $n_h$ total heads, the generating function takes the form: 
\begin{equation}
F_{tot} = \sum^N_{i=1}\sum^{n_h}_{h=1}F^{i,h},
\end{equation}
where 
\begin{equation}
F_{i,h} = \log \sum^{i}_{a=1}\exp\left(q_{i,h}\cdot k_{h,a}^{T}+\zeta_{h,i}\cdot v^{T}_{h,a}\right).
\end{equation}
The output projection weight is included in the definition of $v_j$ here, meaning that 
\begin{equation}
v_{b,A} = x_{b,\bar{B}}(W_O W_V)_{A, \bar{B}}
\end{equation}
where $W_O\in \mathbb{R}^{d_{h}}\times \mathbb{R}^{d_{emb}}$ denotes a head size slice of the output projection weight and $\bar{B}\in \{1,\cdots,d_h\}$ spans the intra-head indices.
In the index notation above, the head indices are barred whereas the embedding space indices are unbarred. 
We proceed focusing on the single-head case, as it makes the presentation simpler, and the multi-head generalization is immediate. 
Note that we demonstrate that our energy function approach also can account for safe softmax in Appendix~\ref{app:safe_softmax}

\subsection{Bayesian interpretation}
\label{app:bayesian}

The fact that it is possible to derive the self-attention operation as the minimization of an energy function implies that it is possible to provide a Bayesian gloss on self-attention by identifying a likelihood function and showing that we can obtain the forward pass of the attention block from computing the maximum a posteriori estimate of this likelihood. 

In particular, we propose the following for the log-likelihood function:
\begin{equation}
\Gamma(\zeta,z) = \sum^N_{i=1}\sum^d_{A=1}\bigg(z_{i,A} \zeta_{i,A}-F(\zeta,x)\bigg).
\end{equation}
We denote by $x$ the input to the self-attention block from which we obtain $q,k,v$ from multiplying it by the weights $W_Q,W_K,W_V$ respectively. 
Let us minimize the above with respect to $\zeta$ and $z$ simultaneously:
\begin{eqnarray}
\frac{\partial \Gamma}{\partial \zeta_{i,A}}=0,
\frac{\partial \Gamma}{\partial z_{i,A}} =0.
\end{eqnarray}
These conditions written explicitly read
\begin{equation}
\zeta_{i,A\,*}=0, \,\,\,\,z_{i,A\,*} = \frac{\partial F}{\partial \zeta_{i,A}}. 
\end{equation}
Plugging the first condition into the second leads to the attention forward pass: 
\begin{equation}
z_{i\,*,A} = \frac{\sum^i_{a=1}e^{q_i\cdot k^T_a}v_{a,A}}{\sum^i_{b=1}e^{q_i\cdot k^T_b}}.
\end{equation}
In all, this means we can obtain the gradient w.r.t. $\zeta$ from MAP estimation of the following likelihood: 
\begin{equation}
z_{i\,*,A},\zeta_{i\,*,A}=\textrm{argmax}_{\zeta,z}e^{-\Gamma(\zeta,z)}.
\end{equation}
Moreover, such a procedure enables us to identify the energy-based model associated with the self-attention function.

\subsection{More Performances Results with a LLama Transformer Model}
\label{app:llama_perf}

To extend our work in section \ref{sec:llama_perf}, and to demonstrate that Tree Attention can be successfully applied to a range of hardware setups, we also experiment with running Llama3.2-1B on a dual NVIDIA RTX 4090 setup. The two 4090s are connected via PCIe networking. Even in this case, we observe a significant 4x speedup (growing to 5x at longer sequence lengths) of Tree Attention over Ring Attention for autoregressive decoding.

\begin{table}[ht]
\centering
\caption{Average Decoding Time (in seconds) comparisons with prefill stage using the 1B Llama 3.2 model with \texttt{Tree Attention} (ours) and \texttt{Ring Attention} (SOTA) across various sequence lengths for 4090s. Average results and standard error ($\pm$) are computed using 10 trial runs.}
\begin{tabular}{@{}cccc@{}}
\toprule
\multirow{2}{*}{Sequence Length} & \textbf{Tree Attention} & \textbf{Ring Attention} & Speedup \\ \cmidrule{2-4}
 & Time (s) & Time (s) & \\ \midrule
8000  & 0.34  $\pm$ \small  0.05 & 1.38 $\pm$ \small 0.07 & $\times$4 \\
16000 & 0.58  $\pm$ \small  0.07 & 2.77 $\pm$ \small 0.04 & $\times$5 \\
20000 & 0.74  $\pm$ \small  0.01 & 3.47 $\pm$ \small 0.04 & $\times$5 \\
32000 & 1.01  $\pm$ \small  0.02 & 5.45 $\pm$ \small 0.03 & $\times$5 \\
\bottomrule
\end{tabular}
\label{tab:decoding_times_4090}
\end{table}

\section{Appendix: JAX code}
\label{app:code}

Below is the \textit{tree\_flash\_decode} method. Our full code base is available here: \url{https://anonymous.4open.science/r/tree_attention-7C32}.

\begin{lstlisting}
import jax
from jax import lax
import jax.numpy as jnp
from functools import partial
from jax.sharding import Mesh,NamedSharding, PartitionSpec as P
from jax.experimental import mesh_utils
from jax.experimental.shard_map import shard_map
from flash_attn_jax.flash import _flash_mha_vjp

in_specs=(P(None, None, None, None), P(None, 'i', None, None), P(None, 'i', None, None))
out_specs=P(None, None, None)

@jax.jit
@partial(shard_map, mesh=mesh, in_specs=in_specs,  out_specs=out_specs, check_rep=False)
def tree_flash_decode(q, k, v):
    def flash_num_lse(q, k, v, config=dict(softmax_scale=1.0, is_causal=False, window_size=(-1, -1))):
        tup = _flash_mha_vjp.fwd(q, k, v, config)

        res,lse = tup[1][3],tup[1][4]
        return res,lse

    loc_res, loc_lse = flash_num_lse(q, k, v)
    a_max_global = lax.pmax(loc_lse, axis_name='i')
  
    num_global = lax.psum(loc_res * jnp.exp(loc_lse - a_max_global), axis_name='i')
   
    den_global = lax.psum(jnp.exp(loc_lse - a_max_global), axis_name='i')
    
    return (num_global / den_global)
\end{lstlisting}

The function uses Flash Attention 2 \cite{dao2023flashattention} to compute the local numerator and denominator, both of which are accumulated between devices using an \texttt{Allreduce} (which is what $\textrm{psum}$ and $\textrm{pmax}$ call). NCCL determines in what pattern these results are communicated.

\section{Theorem~\ref{thm:1} Proof}
\label{app:proof}

We prove theorem~\ref{thm:1} below.

\begin{proof}

    \item \textbf{Sequential Case:}
On a single GPU, the reduction operation over an array of size \(N\) has a time complexity of \(O(N)\) since the processor must sequentially process each element.

    \item \textbf{Parallel Processing with \(p\) Processors:}
Divide the array of size \(N\) into \(p\) chunks, each of size \(\frac{N}{p}\).
Each processor performs the reduction operation on its chunk independently. 
The time complexity for each processor is \(O\left(\frac{N}{p}\right)\).

    \item \textbf{Combining Partial Results:}
The partial results from the \(p\) processors need to be combined.
Using a tree pattern for reduction, the partial results can be reduced in \(O(\log p)\) steps. 
Each step involves combining pairs of results, halving the number of results at each step until only one result remains.

    \item \textbf{Total Time Complexity:}
The total time complexity is the sum of the time complexities for processing the chunks and combining the results:
        \[
        O\left(\frac{N}{p}\right) + O(\log p).
        \]

This proves that the time complexity of a reduction involving an associative operation over an array of size \(N\) is \(O\left(\frac{N}{p} + \log p\right)\) when using \(p\) parallel processors, and it reduces to \(O(\log N)\) when the number of processors is equal to the size of the array.
\end{proof}

\section{Computing safe softmax}
\label{app:safe_softmax}

While, mathematically, attention utilizes the softmax operation, in practice this is often numerically unstable using relatively low precision operations.
To address this, a mathematically equivalent function, the `safe softmax' is instead used which subtracts all dot products in the exponential by the max. This ensures that all values being exponentiated are less than 1 and  hence less likely to explode and cause numerical instability.
Here, we demonstrate that our energy function approach also can account for safe softmax.

Let us suppose we compare our generating function
\begin{equation}
F_{tot} = \sum_{i}\log\sum_{a=1}^i\exp\left(q_i\cdot k_a^T+\zeta_a\cdot v^T_a\right)
\end{equation}
and a slightly modified one: 
\begin{equation}
F'_{tot} = \sum_{i}\log\sum_{a=1}^i\exp\left(q_i\cdot k_a^T+\zeta_i\cdot v^T_a-m_i\right).
\end{equation}
When we take the derivative of these two quantities, we see that we get the same result:
\begin{equation}
\frac{\partial F_{tot}}{\partial \zeta_i}\bigg\vert_{\zeta_i=0} = \frac{\partial F'_{tot}}{\partial \zeta_i}\bigg\vert_{\zeta_i=0}.
\end{equation}

To see it explicitly:
\begin{equation}
\frac{\partial F'_{tot}}{\partial \zeta_i}\bigg\vert_{\zeta_i=0} = \frac{\sum^i_{a=1}\exp(q_i\cdot k^{T}_{a}-m_i)v_{a}}{\sum^i_{a=1}\exp(q_i\cdot k^T_{a}-m_i)}\end{equation}\begin{equation*} =\frac{\sum^i_{a=1}\exp(q_i\cdot k^T_{a})v_{a}}{\sum^i_{a=1}\exp(q_i\cdot k^T_{a})}.
\end{equation*}
Normally, when computing the softmax in an online fashion, this procedure is performed where $m_i$ is the row max of $q\cdot k^T$.
This shift makes it so that the sum of exponentials doesn't lead to overflows.

\section{Notations for equations}
\label{app:notations}

Here is a summary of the various variables and indices that will be used in the coming sections:
\begin{equation}
\begin{aligned}
&\text { TABLE I: Variable names. } \\
&\begin{array}{l|ll}
\hline x & \text{Attention Input}
\\q,k,v & \text {Query, key and value vectors } \\ \Gamma & \text {Attention Log-likelihood} \\
\zeta & \text {Source vector } \\ m & \text {Max of $q\cdot k^T$ } \\
Z & \text {Partition function } \\ z & \text {Activation vector } \\
\textrm{n}& \text{Attention numerator}\\
\textrm{d}&\text{Attention denominator}\\
\textrm{lse} &\textrm{Attention score logsumexp}\\
F & \text {Generating function } \\
P & \text{Attention score probability density}&\\
\end{array}
\end{aligned}
\end{equation}

\begin{equation}
\begin{aligned}
&\text { TABLE II: Index names and ranges. } \\
&\begin{array}{l|ll}
\hline N & \text{Sequence length}\\
d & \text{Embedding dimension}\\
d_h & \text{Head dimension}\\
p & \text{Number of devices}\\
t & \text{Chunk size N/p}\\
b & \text{Batch size}
\\a,i,j \in \{1,\cdots,N\} & \text {Sequence Indices} \\ A,B \in \{1,\cdots ,d\} & \text {Embedding indices} \\
\bar{A},\bar{B}\in \{1,\cdots,d_h\} & \text {Intra-head indices } \\ h\in \{1,\cdots, n_h\} & \text {Head indices } \\
\hat{a},\hat{b} \in \{1,\cdots,t\} & \text {Intra chunk indices } \\
\end{array}
\end{aligned}
\end{equation}

\end{document}